\documentclass[nonacm,sigconf]{acmart}
\renewcommand\footnotetextcopyrightpermission[1]{} 

\AtBeginDocument{%
  \providecommand\BibTeX{{%
    \normalfont B\kern-0.5em{\scshape i\kern-0.25em b}\kern-0.8em\TeX}}}

\definecolor{red}{rgb}{0,0,0}

\usepackage{subcaption}

\newcommand\blfootnote[1]{%
  \begingroup
  \renewcommand\thefootnote{}\footnote{#1}%
  \addtocounter{footnote}{-1}%
  \endgroup
}

\begin{document}

\title[Envisioning Narrative Intelligence: A Creative Visual Storytelling Anthology]{Envisioning Narrative Intelligence: A Creative Visual \\ Storytelling Anthology}

\author{Brett A. Halperin}
\orcid{0000-0002-3555-6637}
\affiliation{%
  \institution{University of Washington \\ Human Centered Design \& Engineering}
  \streetaddress{}
  \city{Seattle}
  \state{WA}
  \country{USA}
  \postcode{}}
\email{bhalp@uw.edu}

\author{Stephanie M. Lukin}
\orcid{0000-0001-8761-167X}
\affiliation{%
  \institution{DEVCOM Army Research Laboratory}
  \streetaddress{X}
  \city{Playa Vista}
  \state{CA}
  \country{USA}}
\email{stephanie.m.lukin.civ@army.mil}

\renewcommand{\shortauthors}{Halperin \& Lukin}


\begin{abstract}
\blfootnote{© Brett A. Halperin and Stephanie M. Lukin. ACM. 2023. This is the author's version of the work. It is posted here for your personal use. Not for redistribution. The definitive Version of Record was published in the Proceedings of the 2023 CHI Conference on Human Factors in Computing Systems (CHI '23), April 23--28, 2023, Hamburg, Germany, \url{http://dx.doi.org/10.1145/3544548.3580744}}In this paper, we collect an anthology of 100 visual stories from authors who participated in our systematic creative process of improvised story-building based on image sequences. Following close reading and thematic analysis of our anthology, we present five themes that characterize the variations found in this creative visual storytelling process: (1) Narrating What is in Vision vs. Envisioning; (2) Dynamically Characterizing Entities/Objects; (3) Sensing Experiential Information About the Scenery; (4) Modulating the Mood; (5) Encoding Narrative Biases. In understanding the varied ways that people derive stories from images, we offer considerations for collecting story-driven training data to inform automatic story generation. In correspondence with each theme, we envision narrative intelligence criteria for computational visual storytelling as: creative, reliable, expressive, grounded, and responsible. From these criteria, we discuss how to foreground creative expression, account for biases, and operate in the bounds of visual storyworlds. \end{abstract}

\keywords{Bias, Creativity, Crowdsourcing, Narrative Intelligence, Narrative Systems, Story Generation, Storytelling, Visual Storytelling}

\maketitle

\section{Introduction}
``To teach computers how to generate a story, we need to understand how humans create one'' \cite{10.1145/3453156}. Artificial intelligence has increasingly supported narrative in a range of contexts, reshaping and expanding how the age-old humanistic practice of telling stories takes hold \cite{murray2017hamlet, murray2011inventing,wardrip2009expressive}. While there is much exciting work in the realms of computational narrative intelligence \cite{koller2005michael} ranging from creative visual storytelling with robots \cite{10.1145/3313831.3376258, 10.1145/3411763.3451785, 10.1145/3491102.3501914} to emergent narrative with co-creative interactive storytelling characters \cite{10.1145/3313831.3376331, 10.1145/3337722.3341861, 10.1145/3402942.3402953}, this work takes a step back from the technology itself to better understand the human creative process of storytelling resulting from a controlled elicitation process. In this paper, we set out to, first, examine human intelligence of storytelling without computationally modeling story generation just yet. Our objective is to establish an exemplar of human-authored stories---an anthology---that can inform story-driven data collection for building narrative intelligence to automatically generate \textcolor{red}{visual} stories with computer vision and natural language generation/processing. Our focus is on creative visual storytelling: creatively deriving stories from images. \textcolor{red}{Thus, for the remainder of the paper, we refer to visual stories as stories that are derived from a sequence of images.}

Visual imagery and language have long since complemented each other in visual storytelling. From children's picture books to comics and news articles, this multimedia nexus forms a complementary interplay between imagery and spoken or written language. While audiences often experience stories and pictures together, visual images alone can also operate as starting points---sources of creative inspiration---for authors to write stories \cite{10.1145/3402942.3402953}. Researchers have found that visual thinking \cite{arnheim1954art,arnheim1969visual} and drawing \cite{Angell2015DrawWA} can prompt storytelling from a multitude of perspectives as long as creativity is not disturbed in the process \cite{buckingham2009creative}. 
This affirms how creative writing and visual imagery are interconnected such that stories can be derived from images to culminate in creative visual storytelling.

Creative visual storytelling, at its core, is an exercise of both assigning meaning to images and sharing that meaning with others. The visual stories that emerge from the meaning are more than a description of events or scenery. In practice, they are narrative arcs with characters who harbor hopes and fears amid conflicts with adversaries in potentially endless circumstances. French literary theorist and semiotician Ronald Barthes asks: ``Where does [the story from the image] end? And if it ends, what is there beyond?'' \cite{barthes1977}. This question of assigning and sharing meaning over a sequence of images has been asked by many researchers \cite{barthes1977,huang2016visual,lukin2018pipeline}. Yet, modern works that design systems for automated story generation from images (also called ``visual storytelling'') shy away from these creative leaps, which have the potential to break the barriers that separate images from a larger story world occurring ``off-screen.'' When sharing the constructed meaning, the storyteller has control over how much to pull from the images as source material, and how much to envision or imagine beyond what appears in the images. Will they, or should they, use imagery that may or may not be apparent from the visual language? How close to the image should they stay, and when is it acceptable to stray? The storyteller's biased perceptions of the image shape the story, which can relay stereotypical character archetypes or harmful messages. 

The truth is that storytelling---with or without computers---can be problematic and harmful, but also generative and creative. Storytelling is inextricably entangled with the biases of the storyteller that can either elevate or marginalize certain histories along axes of gender \cite{10.1145/3173574.3174105}, as well as race and social class \cite{10.1145/3529705}. This is in part because there are many ways to tell stories and a plurality of sides to every story. When it comes to telling stories with images specifically, much work points to the different ways that stories can be derived from the same basic sequence of events (e.g., in text \cite{wright1986exercises} or through comic strips \cite{madden99ways}). Part of what drives the variation in spontaneous stories, or storytelling ``in the wild,'' is the unique position and biases of the storyteller particularly in relation to the images. With this in mind, training a system to automatically generate such stories based on human-authored stories will reproduce human biases. On one hand, this has potential to generate narratives that advance harmful prejudices about people. But on the other hand, inevitable biases may be generative when explored critically and creatively \cite{10.1145/3532106.3533449}. Given this paradox, before computerizing stories with potentially hazardous systems (as prior work on algorithmic oppression \cite{noble2018algorithms} and biases in computer-based vision tasks \cite{buolamwini2018gender} have shown), we aim to first understand the nuances and risks that arise when people (never mind machines) undergo a systematic process of deriving stories from image sequences. To investigate this, we set out to assemble a human-authored creative visual storytelling anthology and critically examine the themes within it.

Our central question is: \textit{What are the varied ways in which human authors approach the same systematic creative process of improvised story-building based on image sequences?} To explore this, we collect 100 different stories derived from 20 image sequences (an ordered set of three images) by 73 unique authors who performed the same improvised creative visual storytelling process. Next, we adapt qualitative research methods for narrative knowledge engineering \cite{o2014applying} to conduct thematic analysis \cite{braun2006using, miles1994qualitative} informed by close reading electronic literature \cite{van2003close} and humanistic HCI methods \cite{bardzell2015humanistic}. Following analyses, we identify five themes: (1) \textit{Narrating What is in Vision vs. Envisioning}; (2) \textit{Dynamically Characterizing Entities/Objects}; (3) \textit{Sensing Experiential Information About the Scenery}; (4) \textit{Modulating the Mood}; (5) \textit{Encoding Narrative Biases}. Based on these themes, we \textcolor{red}{envision narrative intelligence criteria for computational visual storytelling} as: creative, reliable, expressive, grounded, and responsible. \textcolor{red}{We discuss how these criteria can guide the development of more equitable training datasets and facilitate the design of robust algorithmic models for visual storytelling.}   

This work makes multiple contributions to HCI literatures on human creative intelligence, visual storytelling, biases, and story generation. First, it introduces a data collection paradigm and curated anthology of the various ways in which crowdsourced authors perform our systematic, creative process of improvised story-building based on image sequences. Secondly, it expands existing debates on visual storytelling and computational models, and the avenues of increasing creative expression beyond literal descriptions and surface-level commentary of images. Thirdly, it exposes how narrative biases are encoded in the process of visual storytelling and offers suggestions for responsibly recognizing them. Fourth, it suggests ways in which plausible storylines can be derived from images by characterizing objects/entities as plot devices and %predicting feasible trajectories to 
\textcolor{red}{sufficiently grounding stories in the visual representations of the storyworld}. Finally, it contributes considerations for narrative intelligence \textcolor{red}{criteria for computational visual storytelling}, based on the findings of our thematic analysis conducted over the anthology. The anthology is available at \url{https://github.com/USArmyResearchLab/ARL-Creative-Visual-Storytelling}.

\section{Related Work}
\label{related-work}

\subsection{Subjectivities in Stories and Systems}

In this paper, we examine not only how visual information can provide creative inspiration, but how it serves as source material in a subjective and improvised process called creative visual storytelling. An example of such interaction between imagery and stories is apparent in the Tell Tale Card game,\footnote{\url{http://blueorangegames.com}} where players analyze pictures, form connections among what they see, and then articulate language to fellow players to tell stories based on how they interpreted the images. While the stories vary player to player and game to game, there are three consistent factors that guide their formation: (1) the environment and presentation in the imagery; (2) the narrative goal of the storyteller; and (3) the audience \cite{lukin2018pipeline}. 
The environment and presentation refer to the content and quality of the imagery, which can spark a range of interpretations depending on the order and speed at which images are presented during the story writing, as well as the mood that the storyteller detects when viewing the images. The narrative goal of the author may be vague and open-ended (e.g., ``to describe the scene'') or narrow and specific (e.g., ``describe an epic adventure''), and spark different choices of what is noticed and narrated
\cite{lukin2018pipeline,thorne2003telling}. Lastly, the role of the audience can shape the oral, written, or multimedia narration style, influencing how delivery is tailored to suit audience personalities \cite{thorne1987press} and contexts in which the stories are encountered. Creative visual storytelling is enacted along variations of these three dimensions.

The cognitive processes behind forming the varied interpretations of creative visual storytelling have been explored by analyzing and eliciting stories told from images. In Barthes' study of an advertising image for {\it Panzani} brand products, which pictured  packets of pasta, tomatoes, onions, and other food items emerging from an open string bag, he deconstructed the story into two types of messages: a literal message (denotation) and a symbolic message (connotation) \cite{barthes1977}. The literal message---the visual transfer---is the visible taken at face value (e.g., recognizing a round red object as a ``tomato'' rather than an ``apple''). He refers to these as ``non-coded'' interpretations, and claims ``all that is needed [to understand the literal message] is the knowledge bound up with our perception.'' The symbolic message---the coded interpretation---he claims, is what each part of the image means and what knowledge is drawn from it [Ibid]. However, this analysis only allows for the most likely or plausible interpretation, rather than inviting many interpretations or venues for exploration. In the work of Huang et al. \cite{huang2016visual}, data collection of visual stories is conducted with a similar distinction between the acts of seeing (observing the literal message), and telling (narrating the symbolic message). The compiled corpus called ``VIST'' (VIsual STorytelling) showcases a range of interpretations missing from Barthes' claim of a single symbolic message. These stories, however, do not appear to elicit author-created backstories or off-stage conflicts. The crowdsourced prompting follows an alignment paradigm that asks authors to write a single sentence per each image in a sequence, culminating in stories that veer toward ``objective'' descriptions of the depictions. A pilot data collection conducted by Lukin et al. attempts to overcome the limitations of this image-to-sentence alignment paradigm \cite{lukin2018pipeline}; however, the analysis does not speak to how grounded the stories are in the images or how people's symbolic messages vary.

It is critical that the range of subjective interpretations of images are understood because a single story that suggests a ground truth can obfuscate biases and marginalize alternate perspectives. Not only are there many sides to every story (as the saying goes), but also a multitude of ways to tell the same story, which visual storytelling elucidates \cite{wright1986exercises, madden99ways} in instances like the Tell Tale Card Game. More broadly, HCI scholars have shown \textit{how} stories are told and {\it whose} stories are told in the visual language of textiles, for instance, can either work to elevate marginalized histories, or marginalize histories such as racial and socioeconomic origins \cite{10.1145/3529705}, as well as gendered legacies \cite{10.1145/3173574.3174105}. Hence, to reorient the false sense of ``objectivity'' in dominant HCI methods, Rosner posits the concept of ``critical fabulations'' as narrative tactics for reworking design \cite{rosner2018critical} and its histories, while also working with Benabdallah et al. to recognize the generative potential of harnessing biases in creative, yet critical ways \cite{10.1145/3532106.3533449}. With that said, irresponsibly imbuing human biases in algorithms can harmfully reproduce and scale them \cite{bolukbasi2016man,kiritchenko2018examining}, particularly along the axes of race and gender surfaced in computer-based vision tasks \cite{buolamwini2018gender}. To address this, interventions have been proposed such as sampling and comparing training data against an ethical checklist \cite{beretta2021detecting,smiley2017say}. While these precautions can reduce execution risks (e.g., generating hate speech \cite{lee2016tay, wolf2017we, curry2018metoo}), deriving stories from images introduces layers of bias that are intrinsic to the visual information that is included or excluded.  

Much work on visual storytelling includes the same kinds of images, which presents opportunities to derive other kinds of stories from otherwise excluded images. Many visual narrative and visual understanding datasets utilize the very same or similar high-quality canonical images of vibrant scenary from image hosting websites such as Flickr (as VIST uses) \cite{everingham2010pascal, plummer2015flickr30k,rashtchian2010collecting,hodosh2013framing,lin2014microsoft,chen2015microsoft,krishna2017visual,ferraro2015survey}. Excluded from analyses are more obscure, low quality, low-resource domains, and automated photographs that are not so ``picture perfect.'' This unexplored aspect of ``environment and presentation of imagery'' \cite{gordon2018playing, lukin2018pipeline} may not only play a role in what kinds of stories are told, but also affect the elicitation of creativity and ``interestingness'' \cite{10.1145/3453156}. In our study to follow, we investigate the implications of using such images to elicit many different stories that showcase a range of connections and interpretations formed in our improvised story-building process. Thus, we manipulate the first factor of creative visual storytelling---the ``environment and presentation'' \cite{lukin2018pipeline}---and allow for the narrative goal and audience to be interpreted by the storytellers, thereby inviting various forms of creative expression.

\subsection{Computational Image-to-Text Models and Visual Storytelling}

HCI scholars have studied computational models of visual storytelling, ranging from interactive storytelling to automatic story generation. While we focus on the latter, we note that researchers have assembled intricate interactive systems where people can co-create stories with virtual characters (e.g., in visually immersive simulation games) \cite{10.1145/3402942.3402953,10.1145/3337722.3341861,10.1145/3313831.3376331}, as well as with virtual agents to facilitate collaborative storytelling in contexts including education, advocacy, and play) \cite{10.1145/3325480.3325488,karimi2018creative,Halperin2023,10.1145/3313831.3376258}. For example, Zhang et al. have explored how AI can support children in creative visual storytelling by facilitating co-creative processes such as drawing, ideating, and story writing \cite{10.1145/3313831.3376331,10.1145/3411763.3451785}. These works explore how machines can augment human creativity, but they do not address how machines can learn to be creative storytellers on their own. 

Our interest in machine creativity for storytelling can be understood as ``a subfield of computational creativity where artificial intelligence and psychology intersect to teach computers how to mimic humans' creativity'' \cite{10.1145/3453156}. In this context, creativity is defined as ``the ability to generate novel and valuable ideas'' \cite{aylett2011research}. Prominent applications of creative computational models and narrative intelligence \cite{koller2005michael} in the form of automatic story generation include entertainment \cite{aylett2011research,mateas2003integrating,yu2012sequential}, gaming \cite{ammanabrolu2020bringing, hartsook2011toward, valls2013narrative}, education and training \cite{aylett2007fearnot,hartsook2011toward,hermanto2019visual}, and reconnaissance \cite{zook2012automated}. Alhussain et al. have surveyed the textual story generation space, spanning resources, corpora, evaluation methods, and factors of story ``interestingness,'' which feed into applications \cite{10.1145/3453156}. This elucidates the myriad of ways to enact computational creativity. 

That creativity, however, is not always the subject of focus in systems relating to images and natural language. Prior works have primarily focused on generating ``objective'' observations that provide a high-level summary \cite{ordonez2011im2text,vinyals2015show}, or description of events in motion \cite{yatskar2016stating,kumar2018dock,li2020gaia,li2022clip}. This body of work does not tend to generate novel and valuable ideas that would be deemed ``creative" \cite{aylett2011research}. Such approaches alone are not enough to build stories, as they cannot garner knowledge beyond what is contained in the images, or infer creative possibilities around what might happen next. While some work has evaluated narrative plots with ranking systems and multiple choice options to select the next ``best'' sentence in the narrative \cite{roemmele2011choice,hill2015goldilocks,mostafazadeh2016corpus}, such approaches strive for a singular story that foreclose other possible interpretations by optimizing for an ``objective'' truth rather than subjective creative expression. Recognizing this gap, some systems have endeavored to utilize ``objective'' image descriptions as foundational building blocks, rather than focal points, to abet planning creative visual stories. \textcolor{red}{Microsoft's Pix2Story generates a caption from an image, and then, using the caption as context, performs style shifting to transform generic text into a narrative form \cite{pix2story,kiros2015skip}. A conceptually similar approach was utilized by Lukin and Eum, however, they} revealed \textcolor{red}{through human evaluation} the desire for creativity to have ample grounding in and relation to the visuals \cite{lukin2022controllable}. (We reflect on this shortcoming in \S\ref{sec:reliablepredictive}).

Rather than using literal descriptions as building blocks in bottom-up story generation, other works have instead functioned top-down. These approaches rely on crowdsourced sentences written about image sequences, and then train models to replicate the patterns and associations found in the text \cite{park2015expressing,huang2016visual,yu2017hierarchically,wang2018no}. These models, however, are not explainable or controllable in the sense that they demonstrate no understanding of story progression or generation of decisions made. These limitations constrain the creative expression of the models, as intrinsically, they do not know what it means to be ``creative,'' and thus cannot know when to make expressive narrative decisions or appropriately generate creativity. In our study to follow, we explore alternate approaches for such systems.

\section{Methods for Collection and Analyses}

\subsection{Collecting the Anthology}
\label{sec:collecting}

We formulated an experimental protocol to crowdsource an anthology of human-authored stories from a sequence of three images. We defined four facets of creative visual storytelling from a sequence of images: Entity, Scene, Narrative, and Title. 
Our facets were designed with both humans and computers in mind, based on the human-process of image understanding and creative storytelling, as well as a modular methodology that AI technologies (computer vision, natural language generation, natural language processing, and machine learning) can execute. This breakdown into facets is adapted from prior work \cite{barthes1977,lukin2018pipeline,huang2016visual}. The selected images represent not only canonical Flickr images, but also non-canonical image sources of a Search-and-Rescue (SAR) scenario (new ``environment and presentation'' \cite{lukin2018pipeline}). 

Each item in the anthology contains all text written by each author for the four facets that accompanied a given image sequence.\footnote{This data was collected following an approved IRB protocol.}
Table~\ref{fig:main_example} exemplifies one \textcolor{red}{item} in the anthology: \textcolor{red}{a complete set of} author responses for each facet \textcolor{red}{for a particular} image-set. The remainder of this section outlines the facets, crowdsourcing interface, image origins, and anthology statistics. (Figures~\ref{fig:hit-instr}--~\ref{fig:hit-page3} in Appendix~\ref{appendix} show complete screenshots of the instructions and interface).

\subsubsection{Definition of Facets}

\begin{table*}[ht!]
    \centering
    \caption{A single anthology item for one author's facet entries for Flickr Dataset, Image-set 8 (Flickr 8). Image License: CC BY-NC-SA 2.0 \textcopyright wimbledonian.)}
    \small
    \begin{tabular}{p{0.8in}p{1.5in}p{1.5in}p{1.5in}}
            & \includegraphics[width=.22\textwidth]{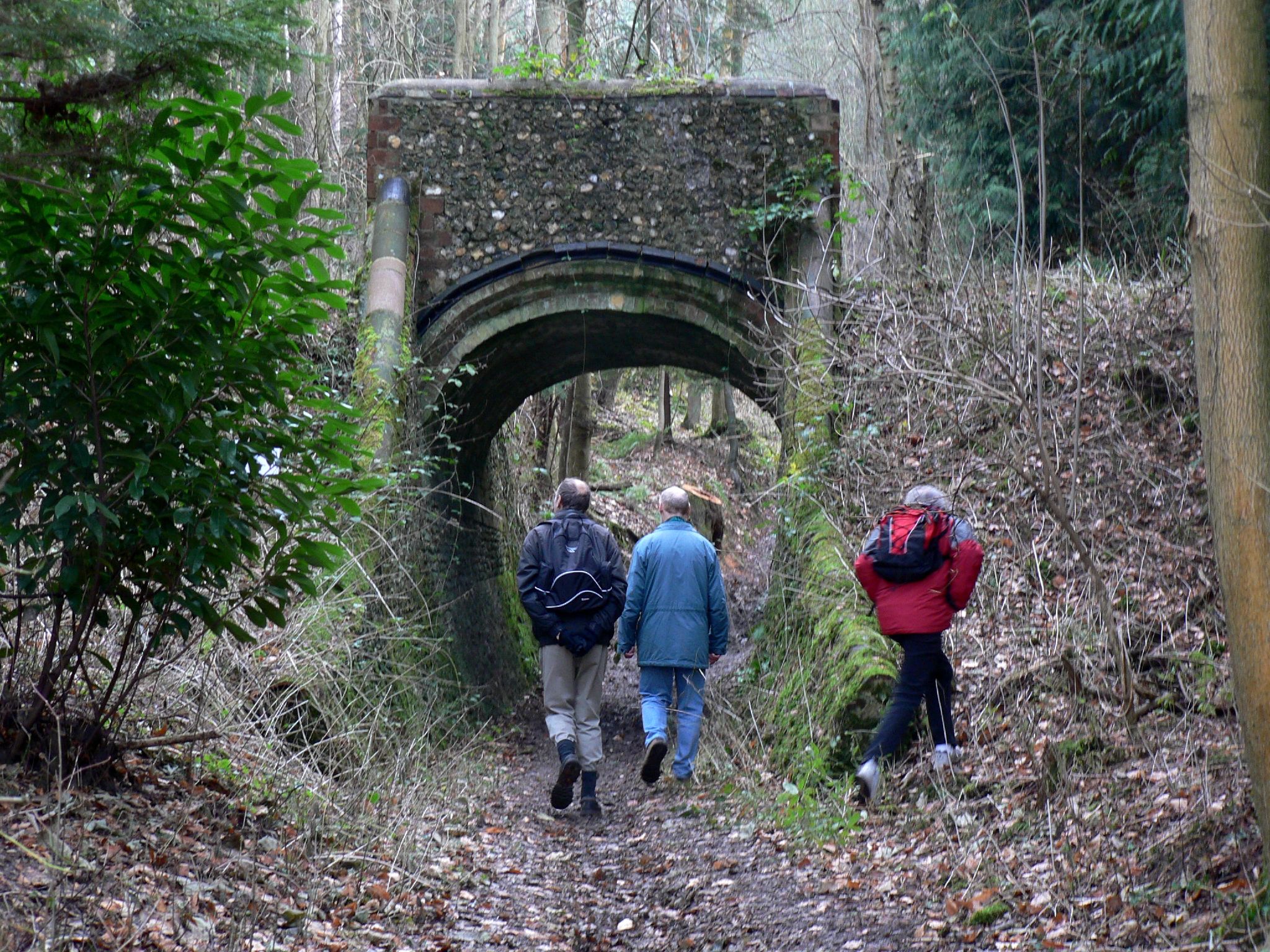}\Description{https://farm1.staticflickr.com/39/82704406\_54de71d38c\_o.jpg}
            & \includegraphics[width=.22\textwidth]{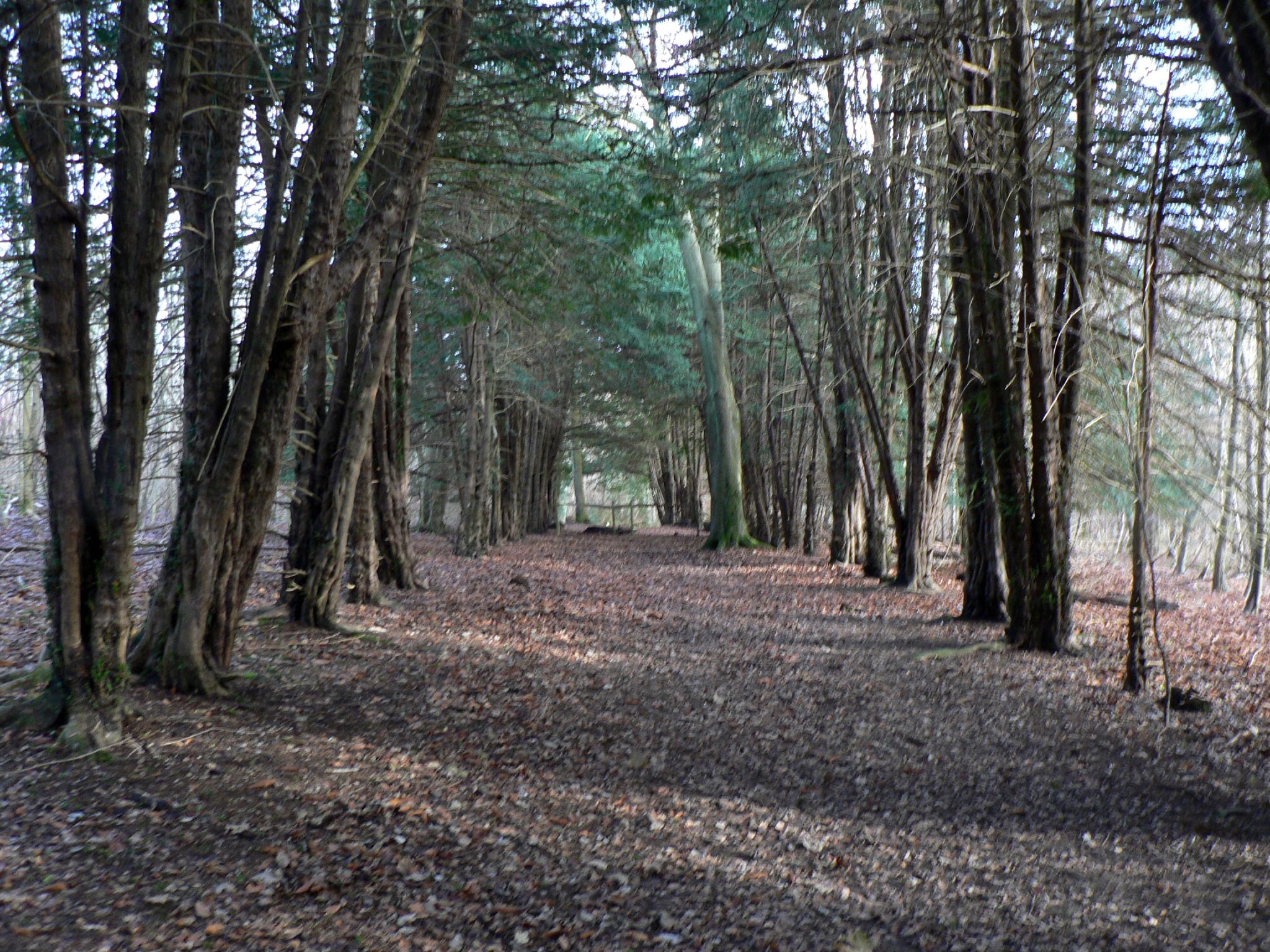}\Description{URL:https://farm1.staticflickr.com/36/82704107\_a6ddd02a5c\_o.jpg}
            & \includegraphics[width=.22\textwidth]{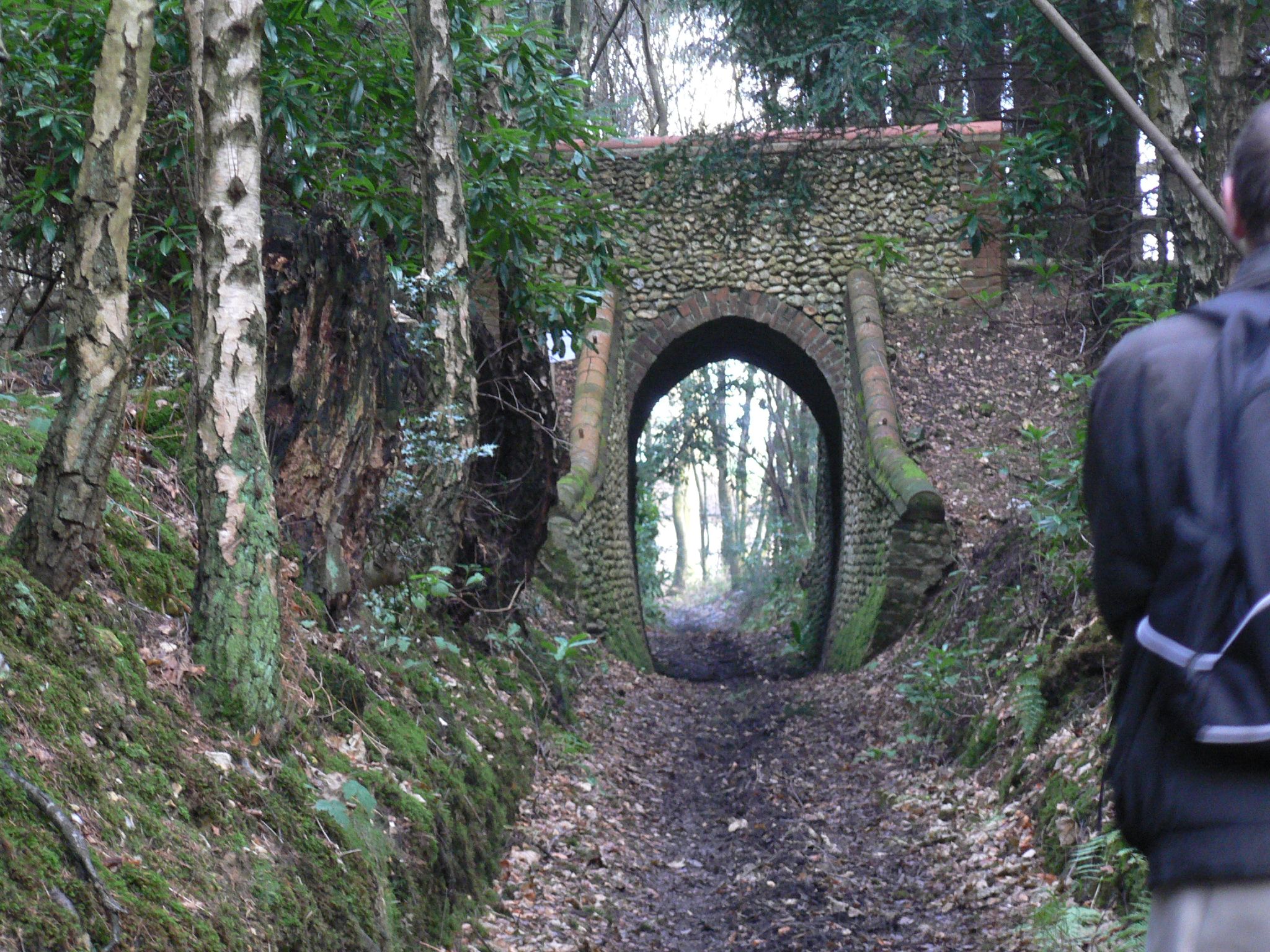}\Description{URL:https://farm1.staticflickr.com/41/82705288\_c08a003c6c\_o.jpg} \\ 
        & \hfil image1 & \hfil image2 & \hfil image3 \\ \midrule
        {\bf Entity Facet} (What Is Here?)
            & hiker in light blue coat \textbackslash{}\textbackslash{} hiker in red coat \textbackslash{}\textbackslash{} hiker in dark blue coat \textbackslash{}\textbackslash{} path \textbackslash{}\textbackslash{} arch \textbackslash{}\textbackslash{} forest \textbackslash{}\textbackslash{} trees \textbackslash{}\textbackslash{} leaves \textbackslash{}\textbackslash{} path \textbackslash{}\textbackslash{} trail \textbackslash{}\textbackslash{} backpacks
            & Trees \textbackslash{}\textbackslash{} Leaves on ground \textbackslash{}\textbackslash{} path \textbackslash{}\textbackslash{} fence \textbackslash{}\textbackslash{} clouds \textbackslash{}\textbackslash{} shadows
            & Arch \textbackslash{}\textbackslash{} path \textbackslash{}\textbackslash{} forest \textbackslash{}\textbackslash{} trail \textbackslash{}\textbackslash{} hiker \textbackslash{}\textbackslash{} backpack \textbackslash{}\textbackslash{} leaves \textbackslash{}\textbackslash{} plants \textbackslash{}\textbackslash{} trees \textbackslash{}\textbackslash{} clouds 
            \\ \midrule
        {\bf Scene Facet} (What Happens?)
            & Three people are hiking on a trail. The trail goes under an arch.  They are in a forest. Leaves on the ground indicate it is Fall. 
            & Trees are casting shadows on the ground.  Leaves cover a broad pathway.  It seems like it is Fall.
            & Hiker approaches an archway. It is a narrow passage.  It is daylight and he is in a forest. \\ \midrule
        {\bf Narrative Facet} (Tell a Story)
            & 
            & Three retirees escape from their care facility.  To make it out before anyone notices they will have to cut thru the forest.  Good thing they dressed warmly.  Just after the arch to path opens up wide and they can see freedom ahead. 
            & Just when they thought the coast was clear another arch appears and they hear rustling in the bushes.  Could this all be a trap?  Will they be sent home? Just then a troll jumps out and says... `I'd like to talk to {\it [sic]} about the extended warranty for your vehicle`. They run like they have never run before. \\ \midrule
        {\bf Title Facet} (Story Title)
            & 
            & 
            & When you least expect it, expect it. \\ \bottomrule
    \end{tabular}
    \label{fig:main_example}
\end{table*}

The first facet was the \textbf{Entity Facet}. Prompting ``what is here?'' facilitated entity/object recognition compatible with how human and computers process images. The facet mimics the visual activation that draws attention when first viewing an image. Authors were asked to list visible entities, describe visual characteristics, and report their degree of confidence in their assessments. Importantly, this facet was not an exhaustive enumeration task.
Authors were asked to list only entities that resonated. This facet was single-image and answered for each image.

The second facet was the \textbf{Scene Facet}. Prompting ``what happens here?'' invited authors to treat the scene as a spatiotemporal snapshot, and provide information about depicted actors and actions.
This facet was akin to a literal explanation of the image, and was not intended to read like a story, but rather a factual description, identifying the location, time, and activities. 
Following a cumulative process, this facet built off authors' prior knowledge about already identified entities to facilitate piecing together the scene. This was also a single-image facet asked for each image.

The third facet was the \textbf{Narrative Facet}. Prompting ``tell a story'' invited authors to narrate across a subset of images following prior responses. Integrating the other facets, this entailed weaving together a span of events and evoking a temporal arrangement of scenes, as well as subjective evaluations and orientations. By inviting leaps into possibilities beyond visual observation, this facet also evoked creativity. 
Unlike the prior facets, this one was multi-image to facilitate storytelling over an image-set sequence. This facet also evoked a sense of natural progression in improvised story-building. Authors did not have all the visual information available to them when starting the story. As images were revealed one at a time, authors were encouraged to adapt their stories  and draw new connections between what they had previously written and what they observed in subsequent images. 

The fourth facet was the \textbf{Title Facet}, asking authors to title their stories. With no requirements, it could be a word, description, or creative expression. This multi-image facet was open-ended, but also implicitly encouraged synthesis.

\textcolor{red}{These four facets were structured to subdivide the visual story-building process into two observable close-ended facets (Entity and Scene), and two imaginative open-ended ones (Narrative and Title). Completing the Entity and Scene Facets yielded observations that could be verified for accuracy by viewing the images. Meanwhile, the Narrative and Title facets could not be systematically mapped to image pixels because they encompassed something greater: plot, conflict, perspective, and characters with complex traits and desires. Each facet built incrementally upon the prior, and encouraged an examination that moved from image assessment to story construction and delivery. Thus, the four facets comprise two tightly bounded facets and two loosely bounded facets to facilitate a balanced process for eliciting visual stories.}

\textcolor{red}{Breaking story-building down into facets builds in part upon prior work, but with key differences. Barthes' literal message \cite{barthes1977} maps well to our Entity Facet. However, his symbolic message combined the description of the image with the narration, a distinction we teased apart with separate Scene and Narrative Facets to disambiguate the observable from the imaginable. Huang et al. \cite{huang2016visual} collected descriptions of individual images (similar to our Scene Facet) and stories across multiple images (conceptually similar to our Narrative Facet, though quite distinct, as discussed further in \S\ref{creativeexpressive}). However, there was no direct relationship between their descriptions and narratives. The narratives were not designed to build upon the descriptions, as our Narrative Facet builds on the Scene. Furthermore, in practice, the authors that Huang et al. recruited to write descriptions and narratives were not the same. One group of authors wrote the descriptions, while another non-overlapping group wrote the stories. In contrast, our paradigm captured the progression from examination to story-building by a single author for a given image sequence (described more in \S\ref{sec:mt}). The closest treatment to this is Lukin et al.'s definition of ``task-modules,'' which mostly correspond to our facets \cite{lukin2018pipeline}, but without a Title Facet that presents an important opportunity for each author to close out, synthesize, and reflect on the entire story.}

\subsubsection{Crowdsourced Writing Interface \textcolor{red}{and Participant Recruitment}}
\label{sec:mt}

The anthology was crowdsourced using a Human Intelligence Task (HIT) on Amazon Mechanical Turk (AMT). The HIT was conducted in stages and advanced through web-pages, where each participant who became an author pressed a button to continue. A new image and facet were introduced on each page to facilitate the sequential writing. They could not edit previous responses or revisit previous pages, although the portion of the narrative that they wrote about the prior images was presented again when completing the narrative for the final image to cross-reference. At least two sentences were required for the Scene Facet and four sentences for the Narrative Facet, \textcolor{red}{which had to be complete, fluent, English sentences. These requirements were explained in the instructions shown to authors prior to them deciding to accept the HIT. Aside from this guidance, storytelling was left open-ended.} (Refer to Figures~\ref{fig:hit-instr}--~\ref{fig:hit-page3} in Appendix~\ref{appendix} for the complete instructions and screenshots of the interface). Three sample responses were supplied to show potential and diverse responses. Participants were encouraged to ``be creative and have fun,'' \textcolor{red}{and were} compensated based on pre-pilot testing of 10-15 minutes per HIT at \$2.50 a HIT: a \$10-\$15 hourly rate. 

\textcolor{red}{Author participation was constrained to those who had at least 100 HITs approved for other AMT Requesters with locations limited to Australia, Canada, UK, and US to reach communities with English fluency. The HITs were released in daily batches and author responses were manually reviewed on a nightly basis before releasing subsequent batches. If authors did not follow the instructions that stipulated to write in complete, fluent, English sentences, then their HIT was rejected and they were added to a ``disqualified'' list that barred them from participating in future batches. Rejections were not issued for any other reason (e.g., due to the caliber of the story). All authors were asked to complete a demographic survey, however, completion could not be enforced on AMT without stymieing data collection. Consequently, we do not have this metadata to report (a source of reflection revisited in \S\ref{sec:bias}).}

\begin{figure}[ht!]
     \centering
         \includegraphics[width=.47\textwidth]{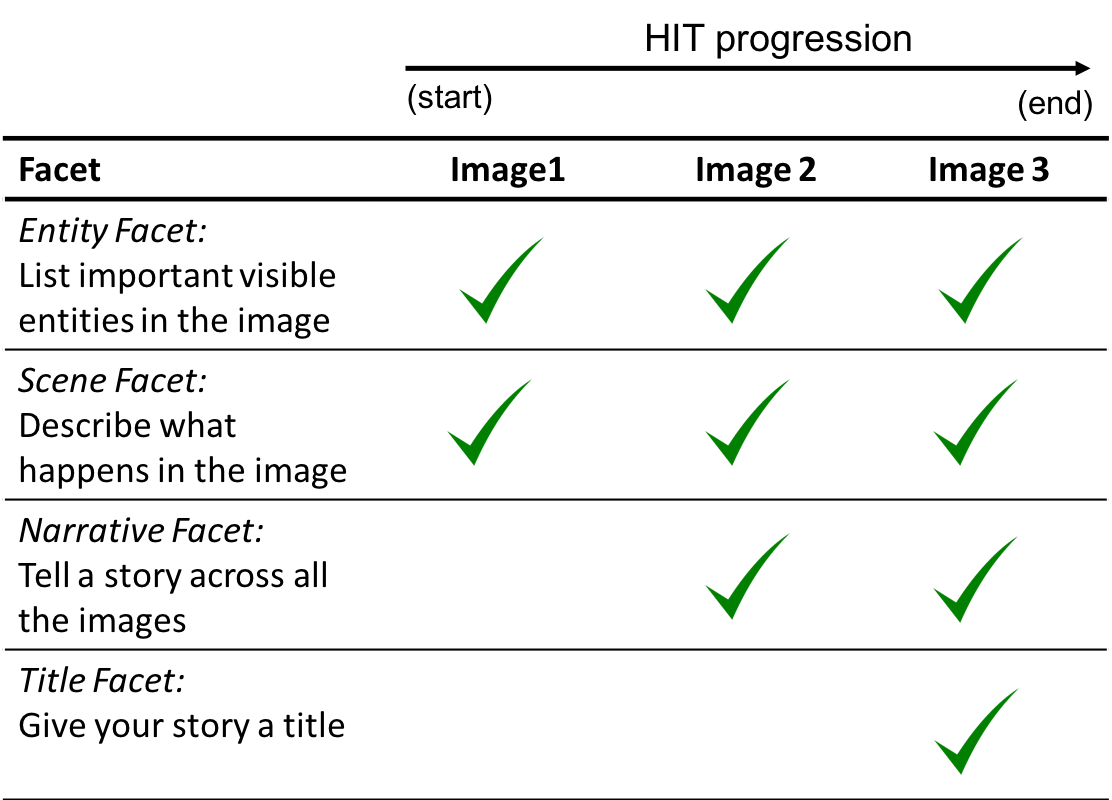}
         \Description{Chart mapping the ``Human Intelligence Task (HIT) progression.'' The four column titles are: Facet, Image1, Image2, Image3. The four rows are: Entity Facet (list important visible entities in the image), Scene Facet (describe what happens in the image), Narrative Facet (tell a story across all the images), and Title Facet (give your story a title). Image1 shows check marks indicating that the Entity and Scene Facets were completed for it. Image2 shows check marks indicating that the Entity, Scene, and Narrative Facets were done for it. Image3 shows check marks indicating that the Entity, Scene, Narrative, and Title Facets were done for it.}
     \caption{HIT Progression across images and facets.} 
        \label{fig:hit}
\end{figure}

Figure~\ref{fig:hit} shows the order of presenting the facets. This figure was also shown to authors in the HIT, so they knew that the story-building process would be improvised, but not what image would appear next. The first page in the HIT was a single-image (image1 in Figure~\ref{fig:hit}) and two single-image facets: Entity and Scene. Following this web-page, was another one with a new image (image2) and Entity and Scene single-image facets. On this page, image1 and image2 were placed side-by-side with the multi-image Narrative Facet. After this page, a final web-page appeared with a new image (image3) and the Entity and Scene single-image facets again, as well as all three images side-by-side with the multi-image Narrative Facet (plus what authors had written on the prior page for reference). Instructions stipulated to continue writing the stories that they had started based on the prior two images. Lastly, this page had the Title Facet. \textcolor{red}{The set of text written about one image sequence by one author culminates in one item} in the anthology (e.g., Table~\ref{fig:main_example}).

\textcolor{red}{A sequence of three images in particular was chosen because it models the beginning, middle, and end stages of a story, as appears in the Aristotelian dramatic structure: setup, climax, resolution \cite{aristotle}. Further,} our decision to withhold subsequent images from the HIT page as the task progressed was intended to mimic the improvised building of knowledge and scene understanding through a naturally occurring temporal progression. Thus, each author was gradually introduced to new visual stimuli and asked to write \textcolor{red}{and adapt a story in wake of uncertainty about what image was going to appear next. The sequence of three images facilitated this improvisation by setting the stage for a narrative to begin based on image1, expanded on with image2, and then finalized with image3.}

\subsubsection{Image Origin \textcolor{red}{and Selection}}

The first set of images used in collecting the anthology originated from Flickr, under Creative Commons Licenses (attribution has been given in the caption of each image-set). These images are high-quality resolution, depicting everyday scenery and touring. We chose a subset of Huang et al's VIST dataset and downselected their image sequences from five to three images \textcolor{red}{to scaffold the Aristotelian dramatic structure, noting that Huang et al. did not provide justification for using five images \cite{huang2016visual}.} Most photographs did not contain images of people, and images with people and children clearly in focus were removed to protect privacy. While the images are licensed for use, the people in them could not know that they were being shown to researchers or AMT workers, and we wanted to limit personal exposure.

The second source of images came from a Search-and-Rescue (SAR) scenario, \textcolor{red}{that we selected for three reasons. First, the SAR scenario scaffolds the potential for authors to develop a ``quest'' plot, which is one of the seven basic types of plots \cite{booker2004seven}. Second, the SAR data exhibits traits in the ``environment and presentation of imagery'' \cite{gordon2018playing, lukin2018pipeline} that we found to be lacking in existing collections of visual storytelling data: in particular, these images were of low-resolution with dim lighting, taken from an atypical camera angle with a camera positioned onboard a small, ground robot. The scenery was devoid of people and depicted the interior of an unfinished building with several rooms and miscellaneous household items. These images starkly contrast with the ``everyday'' high-resolution circumstances in VIST, and thus contribute to the diversity of data representation. Our third reason for selecting SAR data was that stories about SAR scenarios may be useful and informative in the real world. The SAR images we utilized were from Marge et al. \cite{marge2016applying} and Bonial et al.'s \cite{bonial2017laying} human-robot experimentation,\footnote{The images were obtained through a data sharing agreement. Reproduction in this paper was approved.} in which a human instructed a Wizard-of-Oz robot to navigate through and take pictures of the environment to complete search tasks (e.g., identify and count shoes, or assess if space could serve as an impromptu headquarters or meeting location). Their experimental design created a low-bandwidth situation in which natural language communication served to ground the collaboration, bridging the gap between what the robot saw and its ability to convey that information to its human teammate. Automation of the robot's ability to express visual information through language requires examining the environment and understanding the possible stories that even a vacant and dimly lit space may foster. To uncover possible stories about life-saving circumstances that may be too hazardous for humans to traverse, we used the SAR image-sets in our data collection}. We selected three images in-order from different experimental runs, and similar sequential images were excluded for sake of diversity.

\subsubsection{Anthology Statistics}
\label{sec:stats}

Five different authors performed the HIT for 20 different Flickr and SAR image-sets for a total of 100 \textcolor{red}{items}
in the anthology (Table~\ref{fig:main_example} shows one complete HIT). Since some authors completed a HIT for both image-sets, 73 unique authors participated in total. \textcolor{red}{There are 300 unique Entity and Scene entries (single-image facets completed for each image), 200 unique Narrative entries (multi-image facets performed twice with two and then three images), and 100 unique Title entries (multi-image facets completed for three images). Thus, with each one assigned a title, there are 100 unique stories in the anthology all together}. 
 Table~\ref{tab:statistics} shows the breakdown by dataset.

\begin{table}[ht!]
  \caption{Anthology Statistics (*Seven authors completed the HIT for both Flickr and SAR datasets.)}
  \label{tab:statistics}
  \begin{tabular}{lccc}
    \toprule
    & Flickr  & SAR  & Total\\
    \midrule
    \# Image-Sets (Sequence of 3 Images) & 10 & 10 & 20 \\
    \# Entity Facets (Single-Image) & 150 & 150 & 300 \\
    \# Scene Facets (Single-Image) & 150 & 150 & 300 \\ 
    \# Narrative Facets (Multi-Image) & 100 & 100 & 200 \\
    \# Title Facets (Multi-Image) & 50 & 50 & 100 \\ 
    \textcolor{red}{\# Unique Stories} & \textcolor{red}{50} & \textcolor{red}{50} & \textcolor{red}{100} \\ 
    \# Unique Authors & 38 & 42 & 73* \\ 
  \bottomrule
\end{tabular}
\Description{This is a table that maps out the anthology statistics. The three columns are: Flickr Dataset, SAR Dataset, and Total. The rows are # Image-Sets, Entity Facets, Scene Facets, Narrative Facets, Title Facets, and Unique Authors. The table shows that 10 image-sets were utilized for each dataset (Flickr and SAR). Additionally, Entity 150 facets and 150 Scene Facets were completed for each image-set. Lastly, 50 Narrative and Title Facets were completed for each image-set. 38 unique authors used the Flickr Dataset and 42 unique authors used the SAR dataset; thus, 7 authors overlapped. In total, 100 stories were collected and 73 total authors participated.}
\end{table}

\subsection{Analyses}

We analyzed the anthology through applied qualitative research methods for narrative knowledge engineering \cite{o2014applying} (based on thematic analysis \cite{braun2006using, miles1994qualitative}), as well as close reading \cite{van2003close} as done in humanistic HCI methods \cite{bardzell2015humanistic}. We adapted O’Neill and Riedl’s approach to qualitative coding inspired by thematic analysis \cite{o2014applying}. We treated each \textcolor{red}{item} in the anthology like an interview transcript to identify patterns, while developing and cross-referencing codes through close reading. In the tradition of literary criticism, close reading (new media specifically) here refers to ``minute and patient reading'' to uncover themes, meaning: ambiguities, tensions, and ironies \cite{van2003close}, as well as patterns in diction, figures of speech, acoustic qualities, symbols, images, style, tense, voice, and syntax (as described in humanistic HCI and methods of Explication de Texte) \cite{bardzell2015humanistic}. These methods were conducted by two annotators (the co-authors of this paper) in four rounds of coding. Coding was done through an iterative process of annotating each story, taking note of patterns, and clustering trend-like codes. Each round consisted of reviewing about one fourth of the Narrative Facets along with their Entity, Scene and Title Facets. After each round, each annotator wrote a memo to synthesize emergent codes and then met to discuss them together. Following iterative coding, annotating, and memoing, both annotators met for several hours to discuss all 100 visual stories one-by-one and finalize the anthology themes.

This approach was undertaken on the belief and basis that, in working toward narrative intelligence (AI-supported storytelling) computers are not merely calculators, but rather literary devices that express human subjectivities. Thematic analysis and close reading allowed for paying critical attention to the kinds of stories that computers may someday generate \cite{van2003close}. This entailed gleaning insights intrinsic to the data and not that which exists outside of it (e.g., author personalities or motivations for writing stories). Since computational models require vast amounts of data, paying such critical attention to each input is arduous. However, we argue that it is crucial for addressing how harmful outputs can result from problematic inputs (e.g., biases toward certain groups). In analyzing human-authored stories, we combined these methods to examine subjectivities, nuances, and risks that may otherwise be overlooked in scaling story generation without scrutinizing the training data.

\section{Findings} 

Our thematic analysis of the 100 visual stories in the anthology resulted in five themes that repeatedly appeared and crosscut authors, datasets, and image-sets: (1) \textit{Narrating What is in Vision vs. Envisioning}; (2) \textit{Dynamically Characterizing Entities/Objects}; (3) \textit{Sensing Experiential Information About the Scenery}; (4) \textit{Modulating the Mood}; and (5) \textit{Encoding Narrative Biases}.  \textcolor{red}{These themes can inform criteria to furnish a rubric that may be used throughout the development of computational visual storytelling systems (as further discussed in \S\ref{sec:discussion}}). In the subsections to follow, we present definitions and curated examples from the anthology by referencing each dataset (SAR or Flickr) with the image-set number (1-10) and the author's title.

\subsection{Narrating What is in Vision vs. Envisioning}
\label{finding:narrating}

The first theme that repeatedly cut across stories was this notion of narrating what is {\it in vision} (that which is ocular and appears in the images) versus what is {\it envisioned} (that which is imagined beyond the visible). 
The approaches identified spanned: 
(1) captioning the images, as in, literally describing them; (2) commenting about the images, as in, deducing reasonable inferences; and (3) contriving circumstances that deviate from what appears in the images altogether. Authors varied in how they navigated this tension---some writing entire stories like extended captions, and others fluctuating among all three approaches. To illustrate this tension and its significance, we take a closer look at three stories written by different authors about image sequence Flickr 8 (introduced in Table~\ref{fig:main_example}).

\subsubsection{Captioning a Description}

Many authors adhered to the systematic process of deriving stories from images in a strict sense by writing stories like captions. Here, they described that which was visible in the images without much speculation beyond the optical circumstances. An example of captioning is a sentence that reads: ``The trail was covered with leaves that had fallen from the surrounding trees'' (\textit{Jack’s Favorite Trail}, Flickr 8). This is a literal description of the images without commenting or contriving circumstances beyond what is depicted. With this caption-like approach, we can glean a certain truth about what the images reveal because the narratives are more fact-based and thus objective. Of course each story, nonetheless, is a subjective interpretation, but in this particular sub-theme, the sentences are relatively more `true' in the sense that they can be directly tied back to the objects and entities in sight. 

\subsubsection{Commenting a Deduction}

Taking a more speculative approach, some authors commented on the images by deducing inferences from the context clues of the visuals. In this, they wrote text that may or may not be `true' about what is depicted, yet falls within the realm of possibilities. With respect to the first image in Table~\ref{fig:main_example}, one author writes ``A group of three friends were taking a Sunday hike at a local park'' (\textit{Arch Encounters}, Flickr 8). This comment is deduced from the three people pictured in nature, yet the claim itself is not necessarily `true' or false. It is a reasonable assumption that the people are friends, that they are in a local park, and that it is a Sunday. Yet, it is also plausible that they have another relationship, are in a different location, and it is a different day of the week. Regardless of how accurate this sentence is, it reflects a middle-ground. The sentence is accurate in the fact that there are three people who appear to be outdoors, but it also extrapolates upon that to make inferences, characterizing the scenery with plausible basis. This middle-ground approach is somewhere between a description and a deviation, illuminating the tension, as well as the ambiguity, between narrating what is in vision and envisioning.

\begin{figure*}[ht!]
     \centering
     \begin{subfigure}{0.32\textwidth}
         \centering
         \includegraphics[width=\textwidth]{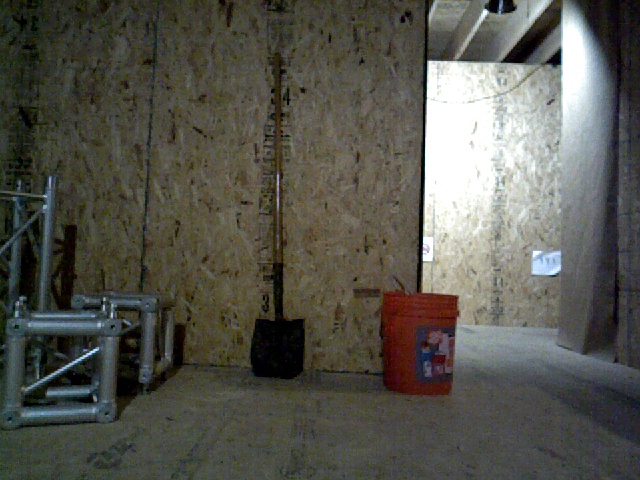}
         \Description{A room in a house with plywood floors and walls, with a shovel and orange bucket on the ground.}
     \end{subfigure}
     \hfill
     \begin{subfigure}{0.32\textwidth}
         \centering
         \includegraphics[width=\textwidth]{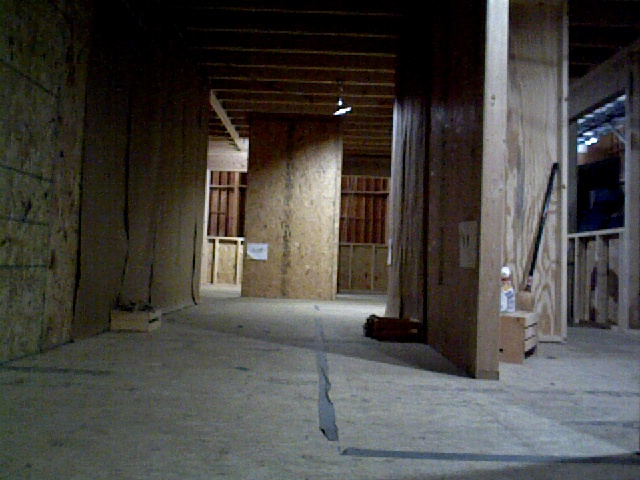}
     \end{subfigure}
     \hfill
     \begin{subfigure}{0.32\textwidth}
         \centering
         \includegraphics[width=\textwidth]{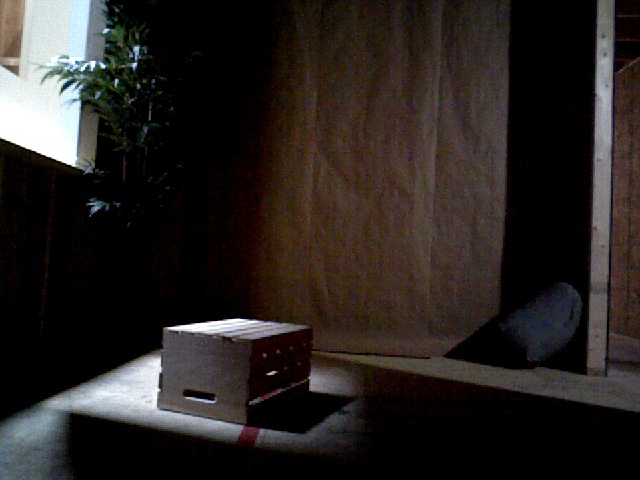}
     \end{subfigure}
     \caption{Search-and-Rescue Dataset, Image-set 5 (SAR 5). Images from Marge et al. and Bonial et al. experiments \cite{marge2016applying,bonial2017laying}.}
        \label{fig:SAR 5}
\end{figure*}

\subsubsection{Contriving a Deviation}

Envisioning circumstances beyond what is depicted, some authors contrived characters, entities, and events that deviated from the images, or could not be known from the presented images alone. While the authors may have drawn creative inspiration from parts of the images (e.g., the scenery), these aspects elucidate the imagination of the author, rather than ground `truth' about the images. As a rich example, one author wrote the following about the same set of images in Table~\ref{fig:main_example}: ``Just when they thought the coast was clear another arch appears and they hear rustling in the bushes. Could this all be a trap? Will they be sent home? Just then a troll jumps out and says... `I'd like to talk about the extended warranty for your vehicle.' They run like they have never run before'' (\textit{When You Least Expect It, Expect It}, Flickr 8). This excerpt showcases a clear deviation from the three people depicted with the introduction of a troll that is nowhere to be found visually. Such instances demonstrate how authors envisioned ways to develop the plot by drawing inspiration from the images, while not adhering to what is strictly in vision. This recurring sub-theme encapsulates how many authors embraced the creative visual storytelling process as a chance to let their imaginations run wild.

\subsection{Dynamically Characterizing Entities/Objects}
\label{finding:dynamically}

This next theme centers on the varying narrative-building approaches authors used to characterize entities/objects in the images. This theme pertains to depicted entities/objects ranging from still-life objects (e.g., a bucket) to more complex entities (e.g., a deer). For purposes of elucidating this theme, to follow is an examination of how five different authors characterized the same object. In Fig.~\ref{fig:SAR 5} is the SAR 5 image sequence explored in this subsection. The subtly depicted plant in the top left corner of the third image is the core object of focus for the analysis that follows.

\subsubsection{The Entity/Object is Overlooked}

The first approach (or lack thereof) to characterizing an entity/object entails overlooking it. For one reason or another, the author did not detect the particular entity/object, or noticed it, but did not narrate it.
There is evidence of a distinction---for example, in listing entities/objects in the Entity Facet, one author did not detect the plant or characterize it in the story (\textit{Voter Fraud Ignored}, SAR 5). Another author, however, listed the plant in the Entity Facet, describing it as a ``tall green plant in the dark corner,'' but then omitted it from the story
(\textit{Scary Voices in the Building}, SAR 5). Both cases of refraining from including the plant in the narration reflect what it means for an author to overlook an entity/object that, nonetheless, appears in the image.

\subsubsection{The Entity/Object is Static}

Some authors characterized entities/objects as static. This approach reflects limitations in how images as static media pose constraints for authors, raising questions around whether a particular entity/object would be characterized the same way if the stories were derived from a video clip instead. Regardless of whether an entity/object is static in reality (e.g., an animal is not actually static, but may appear to be in a snapshot), this approach centers on narrating it as inanimate. One author writes, ``The mother who walked through the home was especially touched when she saw the plant that had been donated'' (\textit{Humanity’s Gift}, SAR 5). Here, the plant is characterized as still-life, and the mother is assigned an active role in viewing the plant, while the plant itself remains passive. This exemplifies how some authors interpreted entities/objects as motionless and narrated them as such, irrespective of whether they may be more dynamic in actuality.

\subsubsection{The Entity/Object is Interactive}
Some authors characterized entities/objects as interactive, where other characters engaged with them in some shape or form. 
For example, with respect to the plant, another author described how a soon-to-be homeowner interacted with it when awaiting for the construction to conclude: ``He placed a plant and a makeshift seat next to a window so that he has a spot he feels is his own'' (\textit{Waiting for My House to Be Built}, SAR 5). This illustrates how the plant was characterized as an object of importance in the story,
revealing the significance of the interaction, while stopping short of the plant having some agency itself and going beyond a static prop. This interactivity demonstrates a tension between static and active that some authors explored.

\subsubsection{The Entity/Object is Active}
To make entities/objects play more integral roles in stories, some authors characterized them as active and dynamic.
For instance, the same author who portrayed the plant as static above, then brought the plant to life, to an extent, by activating it. This also demonstrates how authors can fluctuate among approaches in the course of one story rather than strictly adhere to any one approach. 
The author writes, ``The plant made her actually visualize the home that was to be as she could picture curtains, carpet, and furniture in a room with the plant in the corner.'' (\textit{Humanity’s Gift}, SAR 5). The plant is the active noun: it ``made'' the mother start to imagine her home in new ways. This contrasts with the static portrayal, in which the mother was the one who ``saw'' the plant. Thus, this captures how some authors characterize entitles/objects as dynamic, putting them in the ``driver's seat,'' so to speak, rather than at the whim of other characters who must interact with them in order for them to be activated in the stories.

\subsubsection{The Entity/Object is Personified}

The last approach to characterizing entities/objects was personifying them. This characterization is in tension with portraying entities/objects as static, active, and interactive. The entities/objects were ascribed human characteristics that animated them in meaningful ways. For example, an author describes a construction crew encountering the plant: 

\begin{figure*}[ht!]
     \centering
     \begin{subfigure}{0.32\textwidth}
         \centering
         \includegraphics[width=\textwidth]{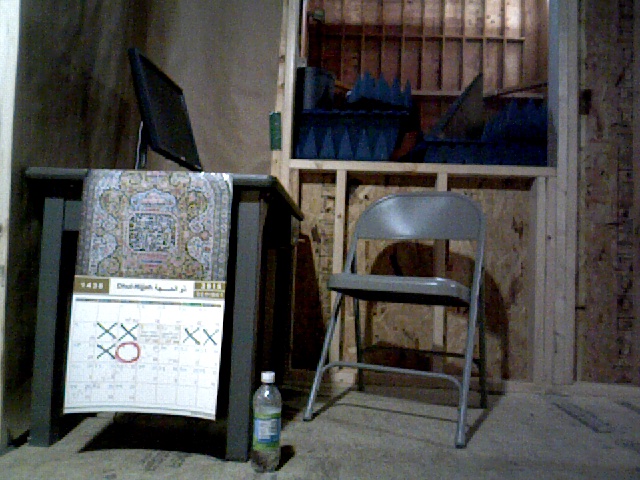}
     \end{subfigure}
     \hfill
     \begin{subfigure}{0.32\textwidth}
         \centering
         \includegraphics[width=\textwidth]{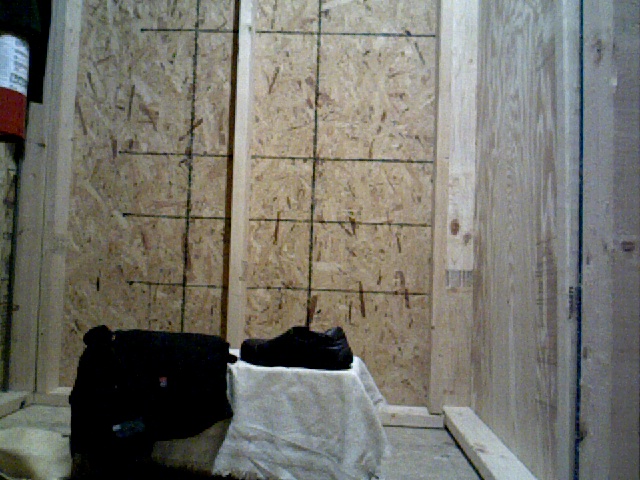}
     \end{subfigure}
     \hfill
     \begin{subfigure}{0.32\textwidth}
         \centering
         \includegraphics[width=\textwidth]{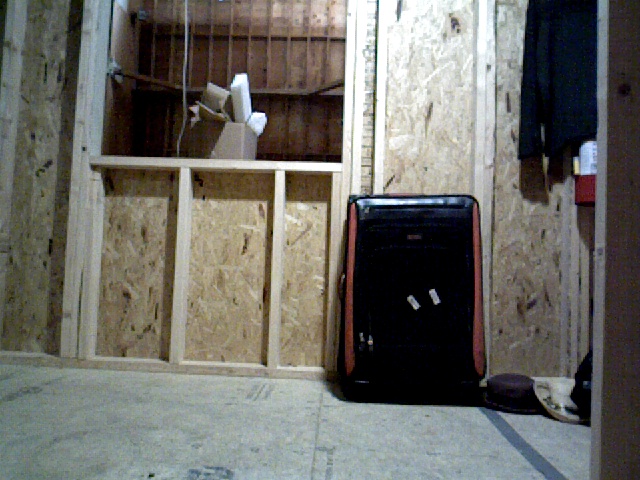}
        \Description{A sequence of images: Image1) an unfinished room with a metal chair, calendar with a red marked circle, and murky water bottle; Image2) another view of the same room with a single black shoe; Image3) another view of the same room with a black suitcase.}
     \end{subfigure}
     \caption{Search-and-Rescue Dataset, Image-set 4 (SAR 4). Images from Marge et al. and Bonial et al. experiments \cite{marge2016applying,bonial2017laying}.}
        \label{fig:SAR 4}
\end{figure*}

\begin{quote}
They have a small indoor tree that they take with them to every building job. It makes the unfinished spaces feel more like home, so that they are more inclined to put forth their best effort. The tree even has a name: Diane. Diane is as much a part of the team as any person. They credit Diane with positive contributions to the team. (\textit{We Couldn't Have Done It Without Diane}, SAR 5). 
\end{quote} This exemplifies how the plant can play a more central role in the story through the personification of it as a character. Furthermore, the plant's stillness is precisely what is dynamic. With the human name ``Diane'' assigned, the plant is portrayed as static and interactive (they ``take'' it with them), yet active in its ``positive contributions to the team'' that emanate from its sheer presence. This approach shows how some authors characterized entities/objects with humanistic traits that heightened their complexity, as well as elucidated the tensions and ambiguities across the sub-themes. 

\subsection{Sensing Experiential Information About the Scenery}
\label{finding:sensing}

Honing in on the sensory information, our third theme revolves around how authors approach the experience of the scenery in two ways. Some authors wrote of a unisensory experience, speaking to the sense of sight when narrating the images, blending the prior two themes---narrating what was in vision alone and characterizing entities/objects as static---in effect. Meanwhile, other authors crafted a multisensory experience, blending the prior two themes again, but this time, envisioning (sensorial) circumstances not in sight, and characterizing entities/objects more dynamically than depicted from what was in sight alone. This includes the sense of sight along with sense of sound, smell, taste, and/or touch. In focusing on the all-encompassing scenery, this theme overlaps with the prior two themes. However, its synergistic effect has distinct implications in terms of experiential information, narrative aesthetics, and plot development. To follow is an examination of the two approaches.

\subsubsection{Sensing a Unisensory Experience of the Scenery}

In sensing only what is in vision and interpreting entitles/objects as static, some aspects of (if not entire) stories expressed unisensory information about the scenery. This is related to, yet distinct from narrating what is in vision for several reasons. For one, this sub-theme has to do with the quantity and quality of experiential information that is conveyed about the scenery, as well as what it reveals about the story. In expressing only what is in sight, the author provides a more limited sense of what it is like to be immersed in the scenery. This synergizes with the characterization of entities/objects as static: visible, but not heard, smelled, tasted, or touched---yet still having the potential to operate as a central plot device.

As an illustration, one author tells about a home remodel project from SAR 4 (Fig.~\ref{fig:SAR 4}), writing, ``We can see that the circled date on the calendar that the project was nowhere near completed at the time expected. Based on the water bottle I feel the calendar represents many months ago. We know that with the timeliness {\it [sic]} that the project was not done in, extreme measures need to be taken'' (\textit{Your Typical DIY’er}, SAR 4). By focusing on the visual details of the objects in the first image, this author used the sense of sight as a narrative device to dramatize a dimension of time. In sensing the sight of the water bottle and what the calendar ``represents,'' the author garners temporality, relaying what the visual marks mean for characters in the scenery. Aesthetically, this narration evokes a sense of sight, but no other senses as the objects are cast as inanimate: the murky water in the bottle is not tasted, the paint or marker chemicals of the red circle are not sniffed, the metal chair is not touched, and sound is not acknowledged. Such instances of unisensory information may develop the plot with visual acuity, as well as provide a sense of what it is like to experience the scenery, but are limited in experience relative to multisensory narrations. 

\begin{figure*}[ht!]
     \centering
     \begin{subfigure}{0.36\textwidth}
         \centering
         \includegraphics[width=\textwidth]{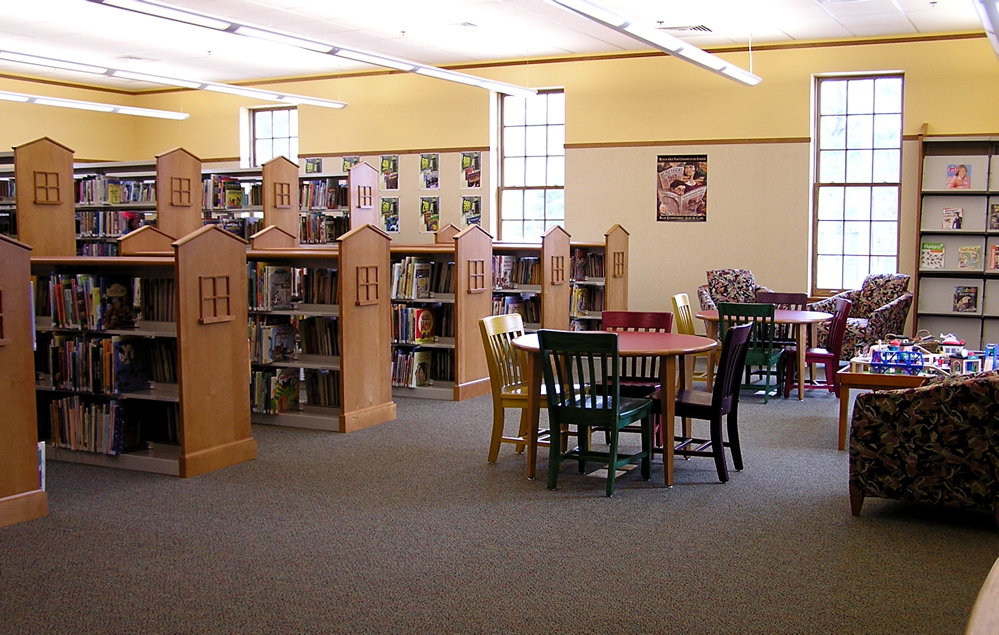}\Description{URL: https://farm1.staticflickr.com/5/7065966_deb4472153_o.jpg}
     \end{subfigure}
     \hfill
     \begin{subfigure}{0.17\textwidth}
         \centering
         \includegraphics[width=\textwidth]{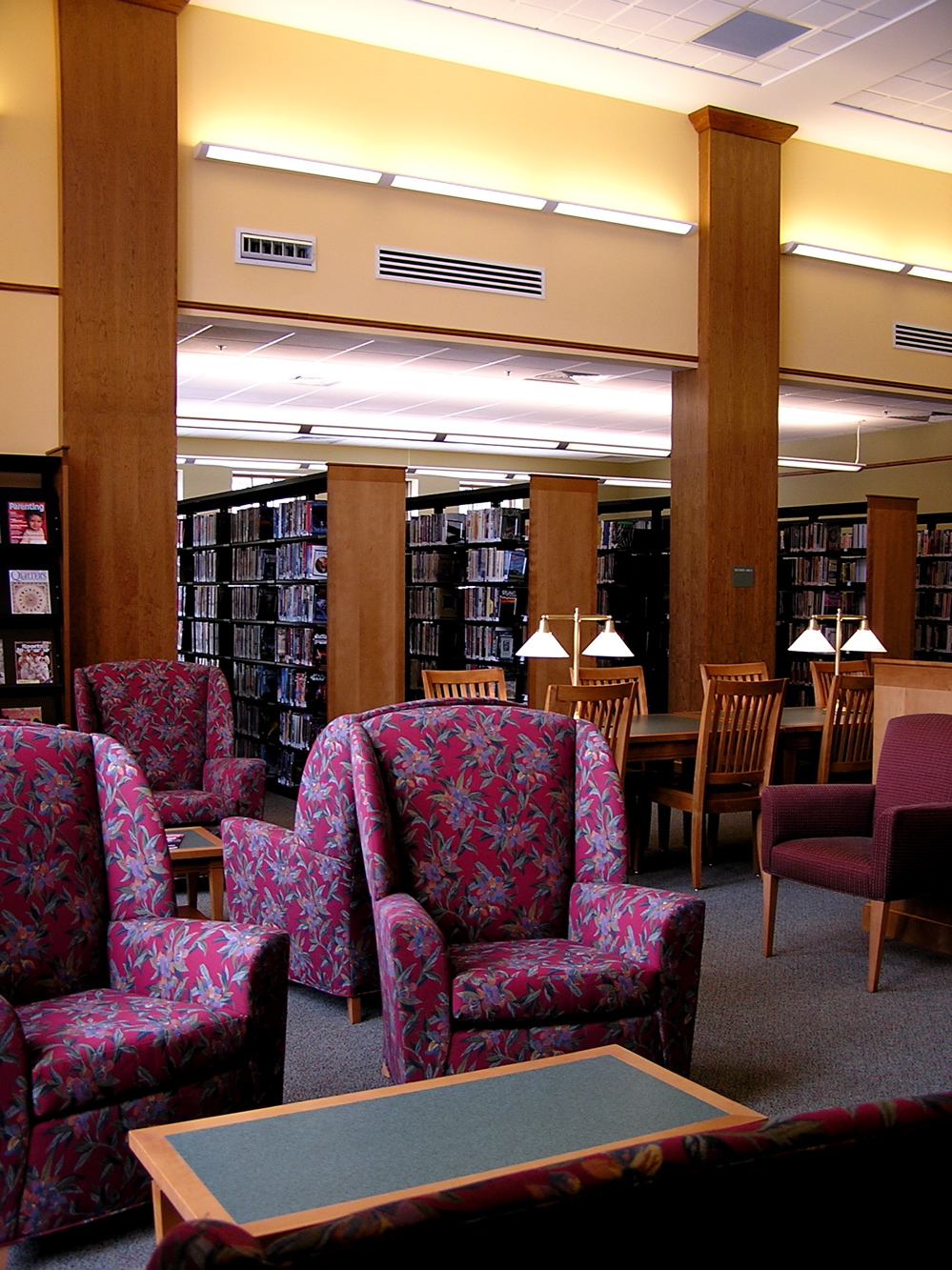}\Description{URL: https://farm1.staticflickr.com/4/7065965_ff5b96cf6a_o.jpg}
     \end{subfigure}
     \hfill
     \begin{subfigure}{0.31\textwidth}
         \centering
         \includegraphics[width=\textwidth]{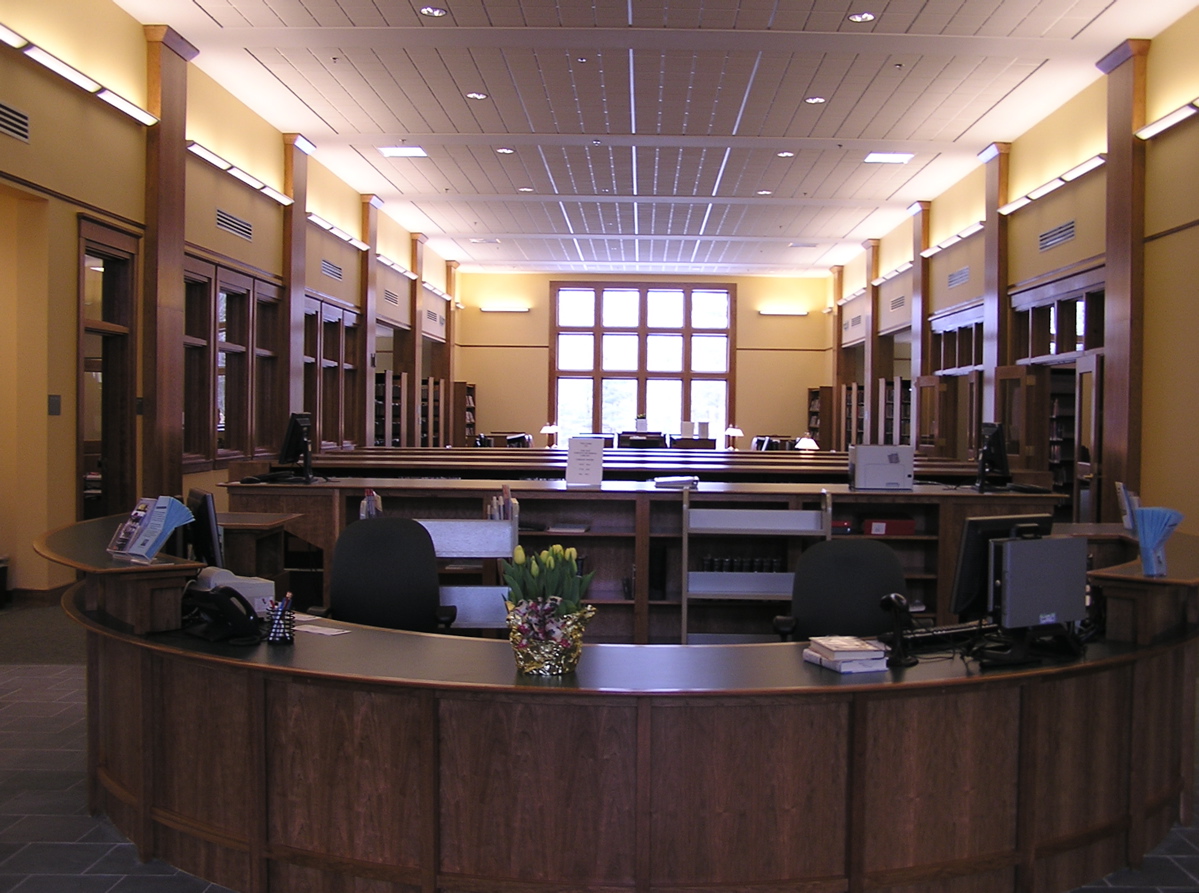}\Description{URL: https://farm1.staticflickr.com/7/7065968_3d62e70095_o.jpg}
    \Description{A sequence of images: Image1) a library with a wooden table, wooden chairs, and wooden shelves with books; Image2) another view of the library, showing two pink, comfy looking chairs with a coffee table; Image3) another view of the library, showing an empty circulation desk}
     \end{subfigure}
     \caption{Flickr Dataset, Image-set 7 (Flickr 7). Image License: CC BY 2.0 \textcopyright calliope}
        \label{fig:Flickr 7}
\end{figure*} 

\subsubsection{Sensing a Multisensory Experience of the Scenery}
In contrast to relaying only a sense of sight, some authors explored more sensoria by conveying greater expressiveness and more experiential information to develop the plot. The multisensory experiences reflect a synergy of envisioning and dynamically characterizing entities/objects. Authors often activated sensoria by imagining how characters traverse the scenery and interact with the entities or objects, eliciting not only a sense of sight, but also smell, sound, taste, or touch. Below is an excerpt of a story from Flickr 7 (Fig.~\ref{fig:Flickr 7}), exemplifying how multisensory narration appears in the anthology: 
\begin{quote}
Walking into the ground floor, I pause to take a deep breath, I love the smell of thousands of books... The wooden columns are the same color as the wood on the bookshelves. The armchairs are the most comfortable chairs I've ever sat in... As I wandered the stacks, the little hairs on the back of my neck began to stand up. Something felt wrong. I sniffed the air again but there were no odors but the smell of old books and newspapers... Then it hit me. There weren't any sounds. Even the quietest library will have the murmur of patrons and librarians... There were no librarians at the desk, no patrons looking at the rows of periodicals... I felt a nervous sweat begin to break out (\textit{Liminal Library}, Flickr 7) 
\end{quote}
This excerpt shows how an author activated multiple senses in expressively narrating an experience of the scenery. The author alluded to the taste of the air (``a deep breath''), the smell of the books, the look of the shelves, the soft touch of armchairs, as well as the lack of sounds and people in sight. This multisensory experience operated as a plot device to build drama around the scenery in effect. The multitude of senses worked together to convey more information, as well as to heighten aesthetics and expressiveness relative to unisensory counterparts. This highlights the decisions of some authors to enrich and advance plots by providing a deeper sense of what it may be like to immerse oneself within the scenary.

\subsection{Modulating the Mood}
\label{finding:modulating}

Some authors further detected a certain mood from the images that they used to shape the plot. Here, mood refers to the aura or ambiance of the environment that is comprised of intangible attributes such as lighting. The diagnosed mood seemed to form a starting point for the narrative arc, and later transformed 
by either intensifying or reversing the trajectory of narrative events. For example, in images across both datasets, many authors identified a ghost-like or gloomy mood that formed the foundation of their narrative. After some development, the authors then forked their plots into one of two different directions. In one, authors pivoted away from the initial mood (e.g., gloomy circumstances) to reverse it and create circumstances that ultimately bloomed. In the other direction, authors leaned into the initial mood (the gloom) such that it intensified and gained momentum until reaching a point of doom. To follow, we unpack these approaches that we coin ``gloom to doom'' (intensifying the mood) and ``gloom to bloom'' (reversing the mood) to demonstrate how authors modulated the mood such that either a worst-case-like or best-case-like scenario unfolded.

\begin{figure*}[ht]
     \centering
     \begin{subfigure}{0.32\textwidth}
         \centering
         \includegraphics[width=\textwidth]{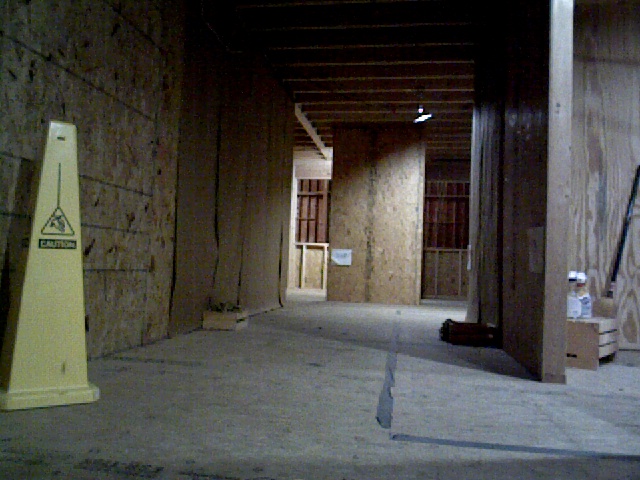}
     \end{subfigure}
     \hfill
     \begin{subfigure}{0.32\textwidth}
         \centering
         \includegraphics[width=\textwidth]{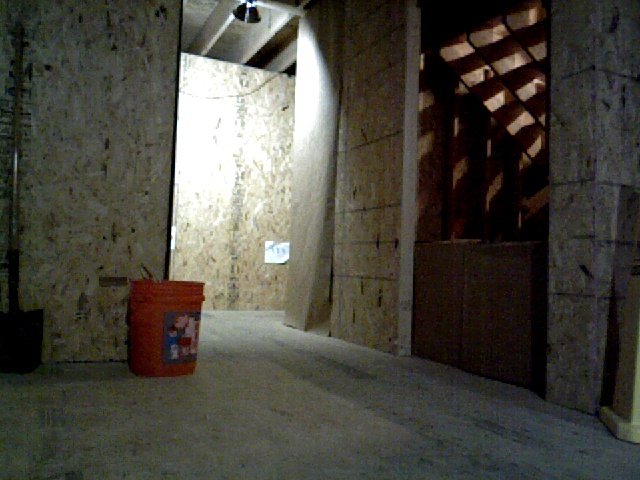}
     \end{subfigure}
     \hfill
     \begin{subfigure}{0.32\textwidth}
         \centering
         \includegraphics[width=\textwidth]{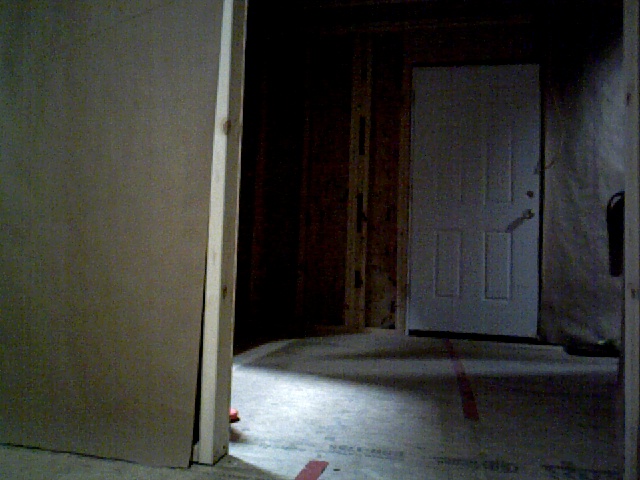}
     \end{subfigure}
     \caption{SAR Dataset, Image-set 1 (SAR 1)  Images from Marge et al. and Bonial et al. experiments \cite{marge2016applying,bonial2017laying}.}
     \Description{A sequence of images: Image1) an unfinished room under construction with a bright yellow caution cone in an otherwise empty room; Image2) another view of the same space under construction with nothing in it except a red bucket and light peering through the otherwise dark and shady room; Image3) another view of the same construction site, but of another room that is even darker and has nothing visible in it except a white door and shadow of light.}
        \label{fig:SAR 1}
\end{figure*}

\begin{figure*}[ht!]
     \centering
     \begin{subfigure}{0.32\textwidth}
         \centering
         \includegraphics[width=\textwidth]{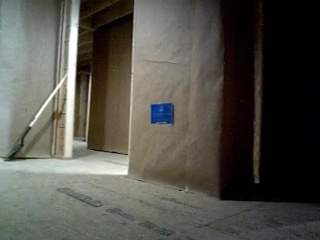}
     \end{subfigure}
     \hfill
     \begin{subfigure}{0.32\textwidth}
         \centering
         \includegraphics[width=\textwidth]{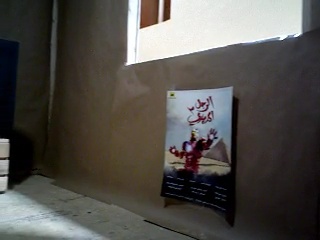}
     \end{subfigure}
     \hfill
     \begin{subfigure}{0.32\textwidth}
         \centering
         \includegraphics[width=\textwidth]{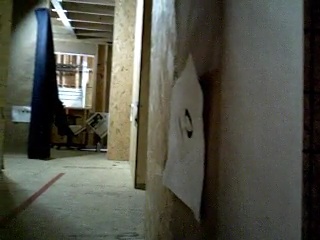}
     \end{subfigure}
     \caption{SAR Dataset, Image-set 10 (SAR 10). Images from Marge et al. and Bonial et al. experiments \cite{marge2016applying,bonial2017laying}.}
    \Description{A sequence of images: Image1) an unfinished room under construction with nothing in it except a blue square on a wall with construction paper on the walls; Image2) another view of the same space with an Iron Man poster and Arabic writing on it; Image3) another view of the same site, but showing an empty room with nothing in it except for an alcove and a curtain on the wall.}
        \label{fig:SAR 10}
\end{figure*}

\subsubsection{Intensifying the Mood}
Some authors established a baseline mood from the images that gradually gained momentum, becoming more intense in the storyline. With respect to the SAR dataset, most authors sensed the depiction of a gloomy mood. In turn, many authors wrote stories with a murder mystery-like genre, reflecting a ``gloom to doom'' scenario, in which the plot became increasingly somber as events unfolded until reaching an ill-fated point for the characters. For example, one author told a story deeming the initial mood depicted in SAR 1 (Fig.~\ref{fig:SAR 1}) as ``drab'' and mapping out a narrative journey that ended with a flesh-eating monster. It reads: 
\begin{quote}
I inherited a home from an aunt... We soon moved in and started to update the home. At this point we just wanted to focus on the main floor and figure out the drab basement later. Our first night in the new home was nice, until we went to bed. It was midnight when I first heard it. We descended down stairs... There was a light coming from the left side of the hall... We went towards the light and there was a door... On the door was a envelope... I opened it and it was from my aunt.
\end{quote}
\begin{quote}
`Behind this door is a great fortune... He will not harm either of you, but in order to obtain any of the money in there, you must feed him each time you enter... You have to feed him meat. Fresh meat. Flesh... Do not be alarmed if he wants human flesh'... The groaning and grinding of a chain across the floor started again just behind the door. (\textit{Inheritance}, SAR 1)
\end{quote} 
Here, the author detected the mood at the onset of the story in terms of intangibles (the ambiance, air, and lighting) and then followed a ``gloom to doom'' trajectory. The author thus modulated the mood such that it escalated until the plot and journey ended with a cliff hanger, suggesting a worst-case-like scenario for the characters.

\subsubsection{Reversing the Mood}
In contrast to intensifying the initial mood, some authors reversed the mood trajectory while developing the plot and journey. With respect to the SAR dataset, many authors initially detected the same ``gloomy'' mood, yet flipped it in the storyline. This transformation encapsulates a ``gloom to bloom'' scenario, in which circumstances started out bleak, but improved to suggest that a best-case-like scenario was on the horizon. For example, of SAR 10 (Fig~\ref{fig:SAR 10}), one author told a story about turning a stark house under construction into a home with tender, love, and care (`TLC'), suggesting a happily-ever-after-like ending, writing:

\begin{quote}
`Everything will be fine,' she says. `It just needs a little TLC,' she says. All the love in the world couldn't make this place feel like home... She has a vision for this place that I can't see. But I will be supportive and positive. The house was filled with junk and trash from the previous occupants. Together, with the use of shovels and tons of trash bags, we were able to clean up really well... I tape my poster on the wall and decided that I'll do my best to give our new home a fair shot (\textit{Faith in the Unseen}, SAR 10). 
\end{quote}
In this story, the gloomy mood initially discerned was modulated such that even a place ``filled with junk and trash'' could transform into a sparkling clean home worthy of ``a fair shot'' for a better life. This suggested a happy ending-like scenario, whereby the reader was left with a sense of closure and even a moral to the story as in the title: have ``faith in the unseen.'' This exemplifies how some authors reversed the initial mood, shaping plots and journeys with transformations that reflect unseen potential, as well as concluding the stories in ways that unfold the possibility of a best-case scenario. 

\subsection{Encoding Narrative Biases}
\label{finding:encoding}

Narrative biases were observed on multiple dimensions: (1) cultural and linguistic biases associated with what authors recognized; (2) the perspective from which authors wrote stories; and (3) the ways that authors cast characters. We unravel these layers of bias below. 

\subsubsection{Linguistic and Cultural Biases}
Based on their familiarity and recognition of what appeared to them in the images, authors encoded linguistic and cultural biases. Linguistic bias refers to how authors subjectively interpreted the appearance of language in the images. Cultural bias refers to the cultural assumptions that authors made about depictions. The Flickr dataset had images depicting German and Arabic language, while the SAR dataset had images with Arabic language. In particular, we focus here on the SAR 10 (Fig.~\ref{fig:SAR 10}) image-set with an image of a poster that has Arabic writing on it. Authors took three different approaches to deriving stories from this image-set. The first approach entailed recognizing the Arabic and then narrating it. For example, one author identified the object as ``a poster of Iron Man in front of ancient Egyptian pyramids with something written in Arabic'' and then told a story deduced from the cultural context, writing ``The construction manager knows that one of the main investors is a Dubai Emirati...'' (\textit{An Update On The Progress At The Construction Site}, SAR 10). This reflects a linguistic and cultural bias, whereby the author knew it was Arabic and then characterized a ``Dubai Emirati'' in the story. 

The aformentioned approach was in contrast to another one: not recognizing the Arabic. Some authors either misinterpreted the Arabic as a scribble or overlooked it entirely. Another author described the same object as ``an iron man poster'' and then narrated it as follows: ``After a long night of drinking, celebration, and forgotten thoughts, they awoke to see an Iron Man poster in the back room'' (\textit{Iron Man Throws a House Warming Party!}, SAR 10). In this case, the author did not mention the Arabic and this omission also reflects particular biases. 

As a third approach, another author expressed uncertainty about the depiction of the poster. The author described the object simply as ``a poster'' and then narrated it as follows: “In another room, somebody had hung some signs. I couldn't understand the language but something was clear...'' (\textit{Signs of Life}, SAR 10). This approach represents a middle-ground sense of recognition relative to the other two approaches. Rather than recognizing, not recognizing, or even making false assumptions about the Arabic, the author narrated a candid sense of confusion in the story itself. As these three different approaches show, authors encoded narrative biases in how they (mis)interpreted the linguistic and cultural contexts.

\begin{figure*}[ht!]
     \centering
     \begin{subfigure}{0.32\textwidth}
         \centering
         \includegraphics[width=\textwidth]{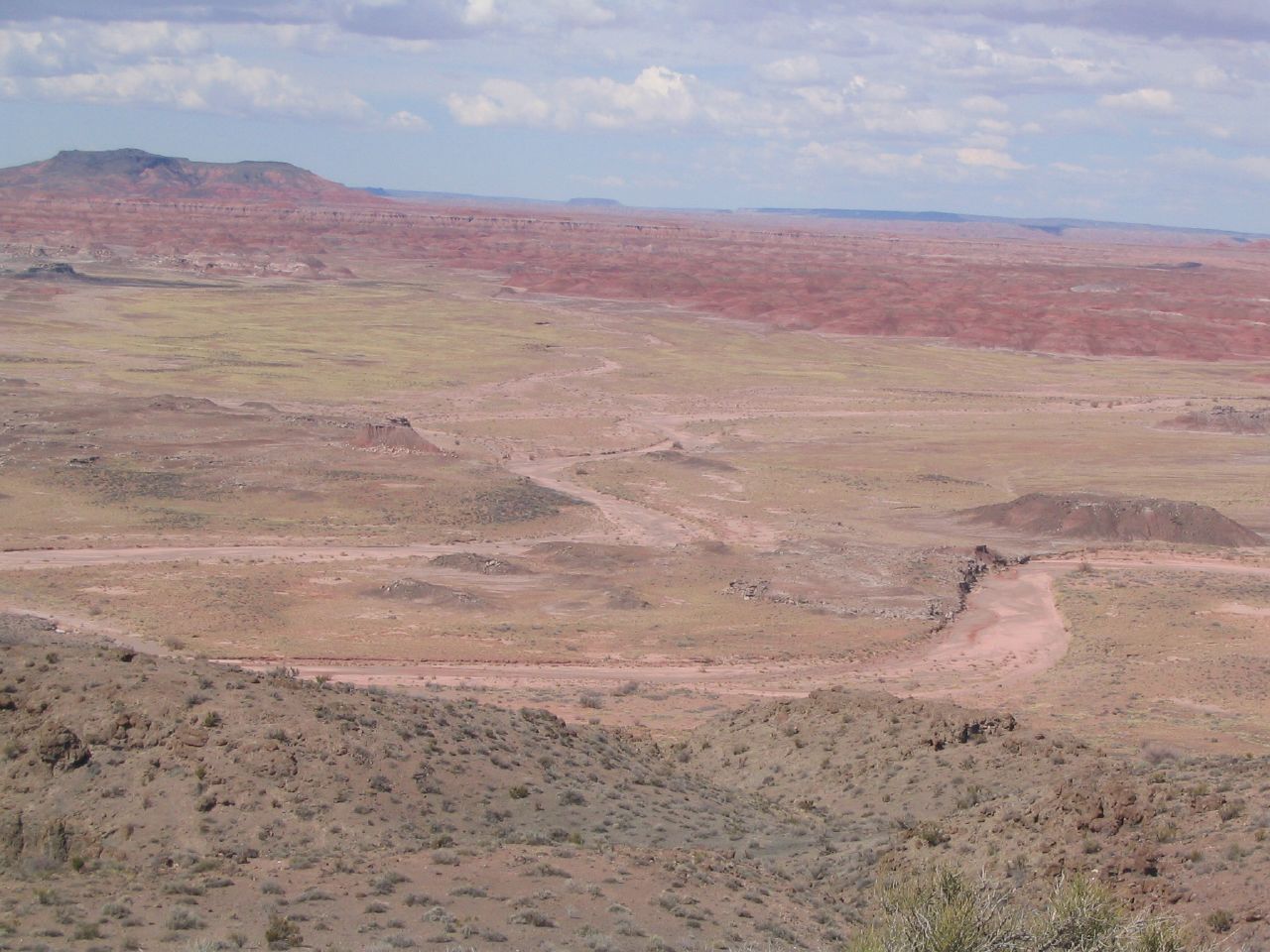}
     \end{subfigure}
     \hfill
     \begin{subfigure}{0.32\textwidth}
         \centering
         \includegraphics[width=\textwidth]{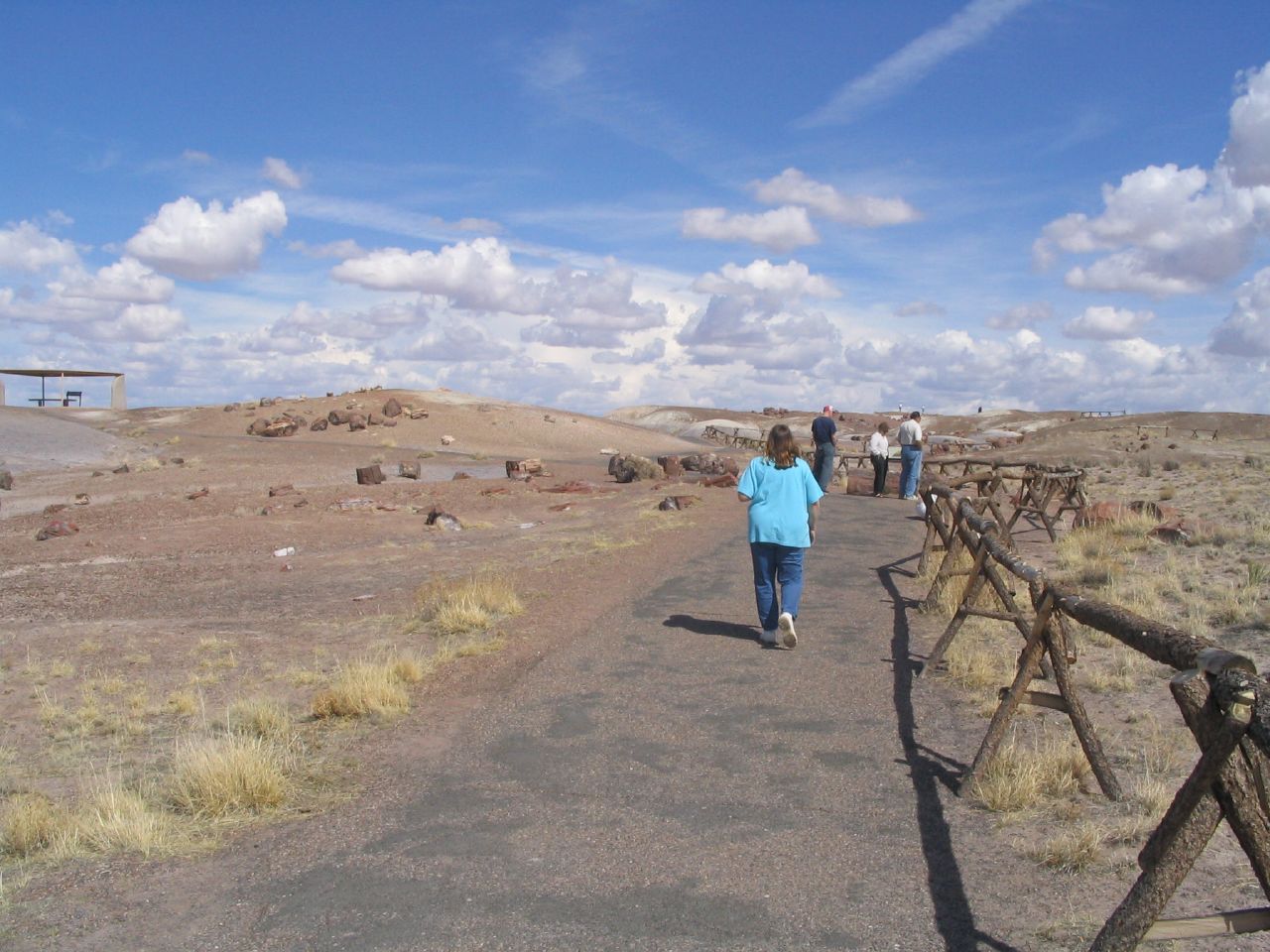}
     \end{subfigure}
     \hfill
     \begin{subfigure}{0.32\textwidth}
         \centering
         \includegraphics[width=\textwidth]{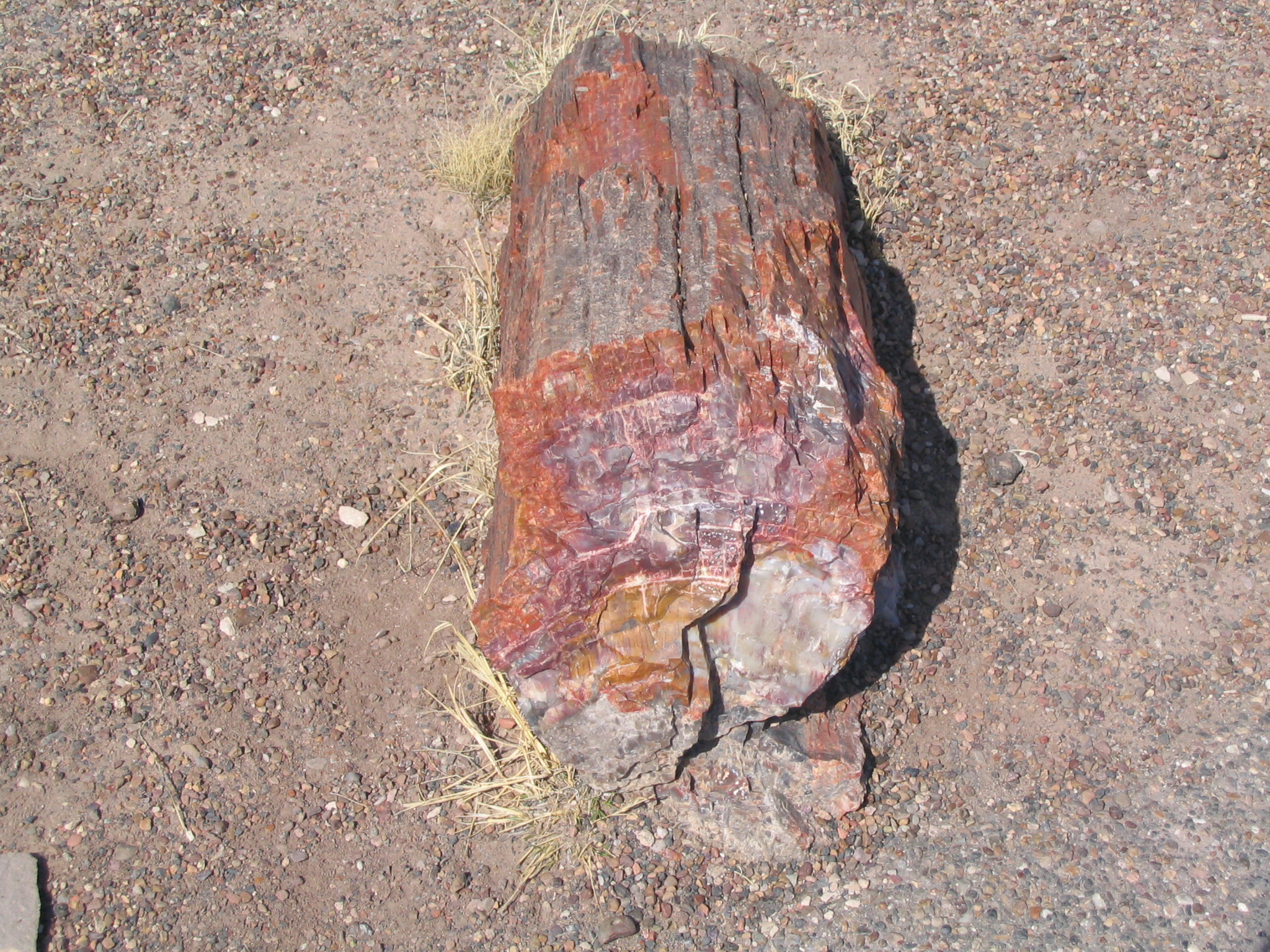}
     \end{subfigure}
     \caption{Flickr Dataset, Image-set 4 (Flickr 4). Image License: CC BY-SA 2.0 \textcopyright kenlund}
     \Description{A sequence of images: Image1) a view of a sandy desert with a blue cloudy sky on a beautiful day; Image2) what looks like another view of the same desert, but with a person in the distance, walking on a path with back turned, wearing a bright blue shirt; Image3) a close up of a stunning gem-like rock that is mostly a burnt orange color}
    \label{fig:Flickr 4}
\end{figure*}

\subsubsection{Narrative Perspective}

The saying `there are at least two sides to every story' rang through in the range of perspectives through which authors narrated. In the SAR dataset depicting a home under construction, most authors fabulated a story from the perspective of construction workers. Meanwhile, few authors wrote from other vantage points such as that of a person facing housing insecurity. The Flickr dataset consisted of a more diverse range of image-sets, which also elicited an array of perspectives. For instance, in a Flickr image-set depicting a train, most authors wrote about the perspective of train riders, whereas just a few wrote about other perspectives (e.g., that of a train worker). Authors narrated varying perspectives, which were sometimes in tension with each other. 

A particularly salient example of how narrative perspective can differ appears in Flickr 4 (Fig.~\ref{fig:Flickr 4}). Two authors interpreted the same image sequence in ways that are reminiscent of two different sides to American history: the perspective of Indigenous people versus that of colonial explorers. For instance, one author fabulated the story below from the perspective of Indigenous people: 

\begin{quote}
The American Indian Sioux tribe ran off all white settlers in a certain region of Arizona for decades, and returned to the traditional native way of life free from interference of the white man. The Sioux headed down into the Red Valley earlier this year to look for more wild American buffalo... As they travelled further on horseback they saw something in the distance... The tribe snuck closer to investigate, and what they saw shocked them. Dozens of white people marching like zombies toward the massive relic... (\textit{Curiosity on Mars}, Flickr 4).\end{quote}
Fabulating the perspective of an American Indian Soux tribe member, this story suggests that the white people do not belong in the scenery---they are invading it. However, this perspective is in contrast to that of another, which narrates the landscape with colonial ascription. The author writes: ```Send in the jets' the world watches as the president makes another attempt to control the situation… `All clear' one of the pilot says... `Down on the ground at Lewis and Clark trail you can clearly see...''' (\textit{Cursed Rock}, Flickr 4) In mentioning a ``president'' and the ``Lewis and Clark trail,'' this story posits a perspective that unequivocally assigns the land to colonial explorers. This perspectives contrasts with how the prior example portrays the land as belonging to Indigenous people. These stories derived from the same image sequence show how narrative biases are encoded in the perspective through which the story is told. 

\subsubsection{Casting of Characters}
Some authors' casting choices reflected social class and gender biases. These biases postured stereotypes, judgmental assumptions, and problematic assertions in harmful tropes. The SAR dataset particularly illuminated questionable casting calls around how characters were presented and represented. The examples we select spotlight narrative biases toward women, working class people, and people facing housing insecurity. To follow is an exploration of instances across image-sets of the SAR dataset, which all depict variations of the same environment.

As many authors interpreted the SAR dataset as a construction site, several cast construction workers in ways that advance harmful attitudes toward working class people. Construction workers, who in the US are often people who have immigrated \cite{kerwin2020us}, were repeatedly cast as not working hard enough. For example, one story, with a biased characterization in the title itself, \textit{The Lazy Workman}, reads as: ``This is a house under construction... A worker has removed his shoes... He is worried his boss, the foreman, might be stopping by today... In secret, he's actually just watching TV and listening for his boss' car. When he hears the boss pull up, he quickly turns off the TV and pretends to be working hard'' (\textit{The Lazy Workman}, SAR 6). Just as this story casts the worker in unfavorable light, another one tells a similar story about a similar SAR image-set. The author describes a contractor slacking off and devising a plan to ``make the client think he's been working'' (\textit{Mario Batali Fan Does Not like to Work}, SAR 8). A construction worker was, yet again, characterized as a ``lazy'' wrongdoer. These casting calls exhibit biases toward working class people with potential to propagate harmful attitudes. 

Similarly, the ways in which authors cast women as characters exhibited gender biases. For instance, an author cast a woman as a manager of the construction site, writing: ``The manager was very happy at the progress of her new room. She was so excited... The workers brought in the Television to be installed... Now if only the manager would let them finish, they might be done by now if she didn't keep asking when it would be done'' (\textit{The New Office}, SAR 7). First, the manager is cast as a woman eager and appreciative of the work, but is then portrayed as incompetent, nagging workers, and impeding progress. In contrast to this unfavorable portrayal, however, another author casts a woman protagonist as competent and handy. She installs drywall, renovates the building herself, and turns it into a leading branch of an organization to rescue an endangered species. The author describes the renovation: ``Tabitha had already decided to start on the entrance area... She can't wait until the remodel is complete. She just knows that, after all of her hard work, her branch of the secret worldwide Save the Turtles club will be among the best in the world.'' (\textit{Tabitha’s Project}, SAR 8). This portrayal contrasts with the gender bias exhibited in the prior example. Together, both examples illustrate how authors encode narrative biases that can either advance harmful attitudes and stereotypes or work to upheave them.

Finally, stories show how authors can encode socioeconomic biases, particularly in casting characters entrapped in poverty, experiencing housing insecurity. As authors drew inspiration from the home-like environment to write storylines about construction, several introduced characters without housing. Rather than portraying these characters as protagonists, however, they were often characterized as antagonistic threats or supporting characters to operate as plot devices for developing more privileged characters' journeys. For instance, one author wrote about a main character, Bob, who encountered an antagonistic squatter who he reports to authority: ``He finally realizes someone is squatting in the building while it's under construction... He calls the building owner to inform them'' (\textit{Bob's Log Stardate 9.28.21}, SAR 7). Rather than casting the squatter as a protagonist to cultivate empathy, the author cast the character experiencing housing insecurity as a criminal that the protagonist ought to help get evicted and possibly even arrested. This bias is further demonstrated in another story that casts people without safe and stable housing as mischievous and thief-like threats. The author writes: ``The workers from a construction company have abandoned an unfinished house... and now the tools that have been left behind are ripe for the picking for any wandering thief... A white door is the only barrier between the outside world and this empty building and its hole where the lock should be ensures easy access... and soon any number of homeless people might discover it'' (\textit{A Job Not Finished}, SAR 1). This author associated ``homeless people'' with thieves, suggesting that the construction workers ought to have safeguarded their property and belongings. These casting calls reflect biases such as contempt for the poor---not character complexity or compassion toward those entrapped in poverty. Such examples spotlight how authors encoded class-based prejudices that can reify harmful rhetoric, attitudes, and tropes, namely those that criminalize poverty, motivate eviction, and vilify those in need. 

\section{Discussion}
\label{sec:discussion}

\begin{table*}[ht!]
  \caption{\textcolor{red}{Computational Visual Storytelling Rubric}}
    \begin{tabular}{lll} 
     \toprule
     \textbf{Criteria} & \textbf{Theme} & \textbf{Phase of Development} \\ 
     \midrule
     \S\ref{creativeexpressive} Creative & \S\ref{finding:narrating} Narrating What is in Vision vs. Envisioning & Data Collection \\ \hline
     %\hline
     \S\ref{creativeexpressive} Expressive & \S\ref{finding:sensing} Sensing Experiential Information About Scenery & Data Collection \\ \hline
     %\hline
     \S\ref{sec:bias} Responsible & \S\ref{finding:encoding} Encoding Narrative Biases & Data Collection \& Labeling \\ \hline
     %\hline
     \S\ref{sec:reliablepredictive} Reliable & \S\ref{finding:dynamically} Dynamically Characterizing Entities/Objects & Visual Story Generation\\ \hline
     %\hline
     \S\ref{sec:reliablepredictive} Grounded & \S\ref{finding:modulating} Modulating the Mood & Visual Story Generation\\
     \bottomrule
    \end{tabular}
\label{rubric}
\end{table*}

Our close read of human-authored visual stories has positioned us close to the data, and therefore able to synthesize our findings into the five themes introduced in the previous section. This literary analysis, which is not commonly conducted when collecting data for large-scale computational applications, reveals the varied ways that authors approached the same improvised story-building process based on image sequences. The first theme, \textit{Narrating What is in Vision vs. Envisioning}, reveals tension among ways of narrating the images by captioning a description, commenting a deduction, or contriving a deviation. The second theme, \textit{Dynamically Characterizing Entities/Objects}, illuminates how people characterize depicted entities and objects as: overlooked, static, interactive, active, or personified. The third theme, \textit{Sensing Experiential Information About the Scenery}, encapsulates how authors narrate the scene to provide either unisensory (solely visual) or multisensory experiential information about it. The fourth theme, \textit{Modulating the Mood}, shows how authors detected a mood in the images and then used that as basis for developing a narrative plot and journey by intensifying it (e.g., a ``gloom to doom'' scenario) or reversing it (e.g., a ``gloom to bloom'' scenario) to culminate in either worst-case or best-case-like scenarios. The fifth theme, \textit{Encoding Narrative Biases}, elucidates how biases permeate the process through cultural and linguistic biases, narrative perspective, and casting of characters.

In correspondence with each theme, we envision narrative intelligence criteria for computational visual storytelling as creative, reliable, expressive, grounded, and responsible. These criteria can serve as a rubric in the development of computational visual storytelling systems, from the collection and labeling of gold-standard datasets, to the generation of stories by training machine learning (ML) algorithms or deep neural networks (as we convey in Table~\ref{rubric}). We operate on the principle that if we ensure the collection of high-quality creative and expressive data, followed by rigorous and responsible labeling and filtering for biases, then the systems trained on such data will be more likely to exhibit desirable attributes than undesirable attributes. This reflects an attempt to avoid the ``garbage in, garbage out'' principle recently revisited by Geiger et al. with respect to modern ML \cite{geiger2021garbage}. 
Using high-quality data and examples, a system can be trained to generate stories from images that exhibit creative, expressive, and responsible properties, that furthermore establish the reliability and groundedness of the systems' ability to generate stories sufficiently related to the input images. We form our discussion to follow as a starting point for computational visual storytelling informed by our close examination of human-authored storytelling. Though we draw attention to these phases of development and relationships specifically, we do not preclude the fact that our criteria may also be consulted at other phases as well. The following subsections describe our call to foreground creative expression (based on the creative and expressive criteria), recognize biases (responsible), and configure the bounds of the depicted storyworlds (reliable and grounded).

\subsection{Creative Expression at the Forefront: Making Visual Storytelling Creative and Expressive}
\label{creativeexpressive}

With our improvised story-building approach, we found that visual storytelling can be creative and expressive. Rather than constraining creative authorial expression \cite{wardrip2009expressive}, our Narrative Facet elicited it. By carving out a distinct place for image description within the Scene facet, the Narrative Facet then encouraged authors to let their imaginations run wild by drawing inspiration from the images but not being restricted by them. We purposely made space for authors to imagine and conceive of possibilities beyond literal extraction of information from the images---to not only caption or comment, but also contrive deviations when appropriate. This entailed utilizing the unisensory visual affordances of the images to their fullest extent by envisioning undepicted circumstances and extrapolating multisensory affordances. We argue that this ability for authors to envision and sense the environment beyond sight was in large part supported by the freedom afforded to them in not only the separation of Scene from Narrative Facet, but also in the writing interface itself. The interface was such that authors could write about {\it any} image that they had previously seen (instead of a strict image-to-sentence alignment) with unlimited sentences to holistically connect the images to a larger experiential storyworld, rather than comment on one image at a time amid syntax constraints. Relative to other approaches for collecting visual story-driven data, ours more so centers and cultivates the creative expression of authors, \textcolor{red}{as exemplified in the narrations that are imaginatively envisioned and richly evocative of multiple senses derived from the image-sets}.

As a point of comparison, we turn to a human-authored story from the VIST dataset \cite{huang2016visual} about Image-set Flickr 8 (Table~\ref{fig:main_example}), which followed an image-to-sentence alignment formula yielding the following: ``It was a perfect day for a hike. [image1] The setting was beautiful and the weather just perfect. [image2] We came across several over passes that were picturesque. [image3]'' (excerpt from VIST data \cite{huang2016visual}). In this paradigm, the authors, also recruited on AMT, saw all images at once rather than in our phased manner that introduced them one-by-one. Showing all images upfront fostered a narrative planning and holistic consideration that was not possible for our authors who had to improvise stories in the wake of uncertainty. What is more, in the VIST example, the story reads like unisensory captioning and commenting. The author was not permitted to write more than one sentence about each image. Even if wanting to contrive an elaborate troll selling car insurance (as one of our authors did for the same image-set), this author had limited space for creative expression in the one sentence restriction. This distinction between asking authors to map one sentence to each image (the VIST approach) and asking authors to map multiple sentences to multiple images (our approach) has implications for eliciting creative and expressive stories. Our Narrative Facet seemed to make more space for contriving multisensory experiences rather than force-fitting a strict paradigm that may be more amenable to training vision and language systems, but more limited in unleashing creative expression, which we argue is important for collecting a range of stories from which to analyze and use as training data. Embracing the creative and expressive criteria in data collection can thus prepare for data labeling and story generation.

\subsection{Responsible Story Collection and Generation: Recognizing Narrative Biases}
\label{sec:bias}
Our findings raise implications and considerations for responsibly collecting and generating \textcolor{red}{visual} stories in the wake of narrative biases. These biases raise questions around the lens through which stories are told at whose cost and whose benefit. \textcolor{red}{To date, biases in visual storytelling in its entirety (the visual and textual combination) have not been examined in human-authored or computational cases. However, in computer vision, a technological subset of image processing,} scholars such as Buolamwini and Gebru have cautioned that algorithmic bias can discriminate along lines of race and gender in \cite{buolamwini2018gender}. \textcolor{red}{Meanwhile}, we similarly found class and gender biases, as well as linguistic and cultural biases. Although we did not find racial bias, we have reason to believe that we would have had we used more images with people. As we found, biases permeate creative visual storytelling across multiple dimensions: language, culture, perspective, and casting characters (e.g., class and gender identities). The potential for advancing harm through storytelling is significant, especially with automated systems that enlarge the scale of it \cite{Halperin2023}.

To address narrative biases in visual storytelling, we do not call for ``solving'' them. Instead, we look to responsibly generate creativity from them, in line with scholars such as Benabdallah et al. whose work on bias makes the case for ``not resolving it but rather tilting it so that it can cast a different light, generate different encounters'' \cite{10.1145/3532106.3533449}. \textcolor{red}{Further, along with Cambo and Gergle, we call for ongoing recognition of biases by adapting concepts of positionality and reflexivity to data science and ML \cite{cambo2022model} for computational visual storytelling in particular}. As our findings show, it is possible to tell different stories about the same image sequence depending on the perspective. Recall how the land in Flickr 4 was either denoted as belonging to the Indigenous people or attributed to colonial explorers (``Lewis and Clark''). This elucidated how every story has many sides to it, and thus a single perspective---whether a person or a machine expresses it---is not the whole story. The side that is told may advance harm along axioms of race, gender, and class as further exemplified by the story motivating the criminalization of poverty and eviction (\textit{A Job Not Finished}, SAR 1). Thus, narrative biases must be examined and questioned: whose side of the story is fabulated, and who and what is at stake in this \textcolor{red}{visual} storytelling? 

To recognize biases tied to inputs and outputs, we offer recommendations for collecting and generating visual stories. Our findings have shown that there is no singular or ground `truth' associated with deriving stories from images. Thus, responsibility ought to entail veering away from attempts at ``objectively'' narrating and generating a singular story for an image sequence. Instead, like how we had multiple authors narrate each image sequence, we advise collecting and outputting a plurality of stories that fabulate diverse perspectives about a given set of images. The inputs also ought to be carefully considered in this process as prior work shows. For instance, a sentiment prediction task with ground truth answers lead Kiritchenko et al. to find that algorithms trained on non-balanced input data exhibit biases towards certain genders or races, unlike those of balanced datasets \cite{kiritchenko2018examining}.
While again we argue that there is no ground `truth' in creative visual storytelling, similar precautions should be taken, nonetheless, to craft diverse inputs.

We delineate these inputs as pertaining to \textit{who} authors stories, \textit{what} images are used, and \textit{what} stories are produced to train visual story systems. Addressing the {\it who} relies on diverse participation in the data collection phase. Since we were not expecting narrative bias to be such a prevalent finding, we crowdsourced authors in a way that ultimately did not support this. While we did ask authors to complete a demographic survey, the data collection platform that we used did not enforce this, and thus we were not able to track this. Moving forward, we invite the community to learn from our reflection and use alternative crowd-sourcing platforms (e.g., Prolific) that are designed for representative data collection. Along with tending to \textit{who} produces training data, responsible visual storytelling is contingent on critical consideration around \textit{what} images are used for story derivation. Our findings suggest the importance of using diverse images around language and culture to assemble training data with proper recognition of entities/objects. Lastly, in terms of \textit{what} stories are collected to someday generate stories with computers, responsibility entails critically examining the inputs before they feed into models. As prior work shows, this is one of the earliest places of intervention to recognize biases in data before ever reaching algorithms \cite{beretta2021detecting} that perpetuate harmful stereotypes \cite{bolukbasi2016man,noble2018algorithms}. In creative visual storytelling, our findings similarly show biases across these dimensions, which call for interventions, including more representative inclusion of sourcing these stories, and marking data as problematic where applicable to serve as negative training examples or to be removed entirely from future datasets. By prioritizing the responsibility criteria, the data collection can be more inclusive and representative of diverse groups. Lastly, the resultant collected data can be annotated and filtered for narrative biases to prevent them from feeding into story generation.

\subsection{Reliable and Grounded Plotting: Operating Within the Bounds of the Visual Storyworld}
\label{sec:reliablepredictive}

When image and text complement one another, they can unite to tell a story that exceeds the sum of its parts \cite{mccloud1998understanding, mccloud1993understanding}---as long as the plot is grounded in the images and the plot devices (entities/objects) are characterized in reliable manners such that they square with the depicted storyworld. Actualizing this synergistic potential of visual storytelling thus requires a coherent---reliable and grounded---relationship between the textual narration and the images. Given what we learned about the varied ways that authors can tell visual stories---coupled with what we already knew about how genre, context, application area, and narrative/author goals affect how stories are told \cite{lukin2018pipeline}---we find that telling visual stories hinges on reliable characterization of entities/objects as plot devices such that they align with literary realism to an extent. Further, we find that the plot trajectory has to be sufficiently grounded in the images. This means that the narrative events do not veer too far off from plausible expectations or what the images depict, but rather align with the representation of the unfolding storyworld without overstepping its boundaries. As our findings show, authors characterized entities/objects within reason and modulated the scenic mood to advance the plot. These sensibilities demonstrate grounding the storylines in the images and reliably navigating the bounds of the visual storyworld, as well as their relations to reality to an extent.

Despite the importance of these sensibilities, we find that existing visual storytelling systems tend to lack reliability and grounding, especially in cases where creative leaps may be possible. To demonstrate this gap, we examine output from Pix2Story, a visual storytelling model developed by Microsoft that generates stories from single images \cite{pix2story,kiros2015skip}. The following is an excerpt generated from an image of a giraffe in front of a harbor: ``There was a ripple in the air, and a large crowd of people began to emerge from the trees and into the sea. They found themselves in a tight spot. There was no sign of life.''\footnote{URL to image:  \url{http://www.cs.toronto.edu/~rkiros/coco_dev/COCO_val2014_000000338607.jpg}. Excerpt from story \#26 from this list of system outputs: \url{http://www.cs.toronto.edu/~rkiros/adv_L.html}} There are two immediate discrepancies between the story and the image. For one, the giraffe prominently standing in front of the boats is overlooked. Secondly, the claim of ``no signs of life'' violates the depiction and tenor of a giraffe's presence, failing to plot events that support the representation of the storyworld. The neural models for text generation that often fuel  computational visual storytelling have been cited to ``write plausible-sounding but incorrect or nonsensical answers'' \cite{chatgpt} (as observed in this example), or to be ``incoherent'' and ``generic'' \cite{holtzman2019curious}. While we again argue that there is no single  right ``answer'' in visual storytelling, if one were to ``plug and play'' these models into a generative application, they would likely fall short in exhibiting properties of human-authored stories, namely the reliable and grounded criteria that make the stories synergistic with the images. Essentially, these criteria can serve as checks and balances on the creative and expressive ones to ensure that the stories do not veer too far off from the visual constraints. Below, we unpack implications of telling stories without these ``guard rails.''

For visual stories to be coherent, entities/objects must be reliably characterized, meaning that they do not overstep the bounds of the storyworld. Achieving this hinges on not only recognizing crucial entities/objects in a scene, but also understanding how they can reasonably operate as plot devices within it. For example, in our close examination of the plant named Diane, we saw that it was personified such that the bounds of reality were pushed, yet kept in check in the storyworld (\textit{We Couldn't Have Done it Without Diane}). The characterization did not break a sense of immersion \cite{murray2017hamlet,murray2011inventing} by exhibiting too outlandish behavior or applying to an ill-suited object. This sophisticated personification raises questions. What characterizations are plausible? Conversely, what kinds of characterizations might exceed the bounds, thereby disrupting immersion? How might stories properly fit into a storyworld that the images represent, but also creatively push the boundaries enough to be interesting? Though our findings surface rather than resolve these questions, we learn that characterizing entities/objects can range from overlooking to activating them. Thus, generating synergistic visual stories requires characterizations that are reliable enough to align with literary realism insofar as that is the genre. 

Where reliable refers to how entities/objects are characterized as plot devices, grounded refers to how the plot trajectory relates to the images. In other words, a visual story that is grounded means that the storyline has \textit{enough} footing in the images to pair together, while at the same time---per our creative and expressive criteria---\textit{enough} imagination to exhibit ``interestingness'' \cite{10.1145/3453156}. This entails shaping the storyline with plot trajectories that can feasibly emerge out of the initial mood detected in the images. As our findings on mood modulation demonstrate, feasible plots can emerge out of an initial mood detected in the depicted storyworld and then reverse or intensify. Our close reading of ``gloom to bloom'' and ``gloom to doom'' stories told from similar visual stimuli reveal this in particular. Such narrative plans, or predictions of feasible story directions, must be grounded \textit{enough} \textcolor{red}{in the images for the story to correspond with them}. However, this raises questions around how to account for a range of possible story arcs grounded in a given image-set, and the perils of quantifying stories to determine this. 

Much story generation has been guided towards the determination that there exists an objective ``best'' storyline---that some are better or more plausible than others \cite{roemmele2011choice,hill2015goldilocks,mostafazadeh2016corpus}. Yet, our analysis demonstrates the myriad of (and potentially endless) possibilities in the emergent narrative space when not constrained to assert a purported singular `truth' about the images. \textcolor{red}{Thus, we call for generative story systems to exhibit} a wide range of plot trajectories by cultivating out-of-the-box thinking such as modulated conceptions derived from the mood. Such thinking may not only intensify the clearly depicted mood, but also cleverly reverse it to make space for a plurality of events and endings rather than optimize for the ``most appropriate'' event or fact-based question and answer. Eliciting a range of feasible plot trajectories---mood modulations grounded in and adapted from images---then entails establishing the bounds of the visual storyworld to therefore design ``guardrails'' for generative models. These guardrails might also serve to ensure the inclusion of critically visible information, for without it, the story diverges too far from the meaning of the image (as the case of the giraffe in the Pix2Story story above). Thus, in generating visual stories, we urge practitioners to use the reliable and grounded criteria tied to the themes. In turn, practitioners can ensure that machine-generated stories have a coherent and sufficient relationship with the images to realize the magnificent synergy of creative visual storytelling.

\section{Conclusion}
\noindent In this study, we collected and analyzed 100 visual stories written about image sequences following a creative visual storytelling paradigm. Our anthology is an exercise of systematically crowd-sourcing improvised and creative stories following the breakout of four facets: Entity, Scenery, Narrative, and Title. Our close reading and analysis revealed five themes that cross-cut authors and image sources to provide insights and discussion about topics so far unexamined in visual storytelling: (1) \textit{Narrating What is in Vision vs. Envisioning} (captioning, commenting, or contriving); (2) \textit{Dynamically Characterizing Entities/Objects} (as overlooked, static, interactive, active, or personified); (3) \textit{Sensing Experiential Information About the Scenery} (as unisensory or multisensory); (4) \textit{Modulating the Mood} (intensifying or reversing it); and (5) \textit{Encoding Narrative Biases} (in culture, language, perspective, and casting characters). In correspondence with each theme, we envision narrative intelligence \textcolor{red}{criteria for computational visual storytelling} as: creative, reliable, expressive, grounded, and responsible. \textcolor{red}{These criteria can support ML practitioners and researchers in} centering creative expression, recognizing insidious biases, and \textcolor{red}{generating coherent visual stories}. 

With these themes and criteria, we plan for future work to bridge the gap between our crowdsourced study and theoretical inquiry. As other theoretical work suggests, our authors may represent a collective, working toward the production of an artifact that is the anthology itself \cite{Korsgaard2022}. While we are not there yet, we conclude that this collective visual story construction lays fruitful groundwork for theory-building. For example, based on our close reading of the collection, we have reason to believe that virtually any author who follows our improvised story-building process will have to confront and navigate---in some shape or form---the tension between narrating what is in vision and envisioning. We plan to pursue this theoretical development further, and release the anthology at \url{https://github.com/USArmyResearchLab/ARL-Creative-Visual-Storytelling} for other researchers to cross-examine and join us in theorizing around.

\begin{acks} % if accepted
We thank our reviewers, as well as Clare Voss and Robert St. Amant for suggesting ways to strengthen this manuscript. We also thank Henrik Korsgaard for discussing the theoretical extensions with us.
\end{acks}

\bibliographystyle{ACM-Reference-Format}
\bibliography{sample-base}


\begin{thebibliography}{88}

%%% the Web address.

\ifx \showCODEN    \undefined \def \showCODEN     #1{\unskip}     \fi
\ifx \showDOI      \undefined \def \showDOI       #1{#1}\fi
\ifx \showISBNx    \undefined \def \showISBNx     #1{\unskip}     \fi
\ifx \showISBNxiii \undefined \def \showISBNxiii  #1{\unskip}     \fi
\ifx \showISSN     \undefined \def \showISSN      #1{\unskip}     \fi
\ifx \showLCCN     \undefined \def \showLCCN      #1{\unskip}     \fi
\ifx \shownote     \undefined \def \shownote      #1{#1}          \fi
\ifx \showarticletitle \undefined \def \showarticletitle #1{#1}   \fi
\ifx \showURL      \undefined \def \showURL       {\relax}        \fi
% invisible to TeX
\providecommand\bibfield[2]{#2}
\providecommand\bibinfo[2]{#2}
\providecommand\natexlab[1]{#1}
\providecommand\showeprint[2][]{arXiv:#2}

\bibitem[Alhussain and Azmi(2021)]%
        {10.1145/3453156}
\bibfield{author}{\bibinfo{person}{Arwa~I. Alhussain} {and} \bibinfo{person}{Aqil~M. Azmi}.} \bibinfo{year}{2021}\natexlab{}.
\newblock \showarticletitle{Automatic Story Generation: A Survey of Approaches}.
\newblock \bibinfo{journal}{\emph{ACM Comput. Surv.}} \bibinfo{volume}{54}, \bibinfo{number}{5}, Article \bibinfo{articleno}{103} (\bibinfo{date}{June} \bibinfo{year}{2021}), \bibinfo{numpages}{38}~pages.
\newblock
\showISSN{0360-0300}
\urldef\tempurl%
\url{https://doi.org/10.1145/3453156}
\showDOI{\tempurl}


\bibitem[Ammanabrolu et~al\mbox{.}(2020)]%
        {ammanabrolu2020bringing}
\bibfield{author}{\bibinfo{person}{Prithviraj Ammanabrolu}, \bibinfo{person}{Wesley Cheung}, \bibinfo{person}{Dan Tu}, \bibinfo{person}{William Broniec}, {and} \bibinfo{person}{Mark Riedl}.} \bibinfo{year}{2020}\natexlab{}.
\newblock \showarticletitle{Bringing stories alive: Generating interactive fiction worlds}. In \bibinfo{booktitle}{\emph{Proceedings of the AAAI Conference on Artificial Intelligence and Interactive Digital Entertainment}}, Vol.~\bibinfo{volume}{16}. \bibinfo{pages}{3--9}.
\newblock


\bibitem[Angell et~al\mbox{.}(2015)]%
        {Angell2015DrawWA}
\bibfield{author}{\bibinfo{person}{Catherine Angell}, \bibinfo{person}{Jo Alexander}, {and} \bibinfo{person}{Jane~A Hunt}.} \bibinfo{year}{2015}\natexlab{}.
\newblock \showarticletitle{‘Draw, write and tell’: A literature review and methodological development on the ‘draw and write’ research method}.
\newblock \bibinfo{journal}{\emph{Journal of Early Childhood Research}}  \bibinfo{volume}{13} (\bibinfo{year}{2015}), \bibinfo{pages}{17 -- 28}.
\newblock


\bibitem[{Aristotle (330 BC)}(1997)]%
        {aristotle}
\bibfield{author}{\bibinfo{person}{{Aristotle (330 BC)}}.} \bibinfo{year}{1997}\natexlab{}.
\newblock \bibinfo{booktitle}{\emph{{The poetics}}}.
\newblock \bibinfo{publisher}{{Dover, New York}}.
\newblock


\bibitem[Arnheim(1954)]%
        {arnheim1954art}
\bibfield{author}{\bibinfo{person}{Rudolf Arnheim}.} \bibinfo{year}{1954}\natexlab{}.
\newblock \bibinfo{booktitle}{\emph{Art and visual perception: A psychology of the creative eye}}.
\newblock \bibinfo{publisher}{Univ of California Press}.
\newblock


\bibitem[Arnheim(1969)]%
        {arnheim1969visual}
\bibfield{author}{\bibinfo{person}{Rudolf Arnheim}.} \bibinfo{year}{1969}\natexlab{}.
\newblock \showarticletitle{Visual Thinking University of California Press}.
\newblock \bibinfo{journal}{\emph{Berkeley and Los Angeles}} (\bibinfo{year}{1969}).
\newblock


\bibitem[Aylett et~al\mbox{.}(2011)]%
        {aylett2011research}
\bibfield{author}{\bibinfo{person}{Ruth Aylett}, \bibinfo{person}{Sandy Louchart}, {and} \bibinfo{person}{Allan Weallans}.} \bibinfo{year}{2011}\natexlab{}.
\newblock \showarticletitle{Research in interactive drama environments, role-play and story-telling}. In \bibinfo{booktitle}{\emph{International Conference on Interactive Digital Storytelling}}. Springer, \bibinfo{pages}{1--12}.
\newblock


\bibitem[Aylett et~al\mbox{.}(2007)]%
        {aylett2007fearnot}
\bibfield{author}{\bibinfo{person}{Ruth Aylett}, \bibinfo{person}{Marco Vala}, \bibinfo{person}{Pedro Sequeira}, {and} \bibinfo{person}{Ana Paiva}.} \bibinfo{year}{2007}\natexlab{}.
\newblock \showarticletitle{Fearnot!--an emergent narrative approach to virtual dramas for anti-bullying education}. In \bibinfo{booktitle}{\emph{International Conference on Virtual Storytelling}}. Springer, \bibinfo{pages}{202--205}.
\newblock


\bibitem[Bardzell and Bardzell(2015)]%
        {bardzell2015humanistic}
\bibfield{author}{\bibinfo{person}{Jeffrey Bardzell} {and} \bibinfo{person}{Shaowen Bardzell}.} \bibinfo{year}{2015}\natexlab{}.
\newblock \showarticletitle{Humanistic HCI and Methods}.
\newblock In \bibinfo{booktitle}{\emph{Humanistic HCI}}. \bibinfo{publisher}{Springer}, \bibinfo{pages}{33--64}.
\newblock


\bibitem[Barthes(1977)]%
        {barthes1977}
\bibfield{author}{\bibinfo{person}{Roland Barthes}.} \bibinfo{year}{1977}\natexlab{}.
\newblock \showarticletitle{Rhetoric of the Image}.
\newblock \bibinfo{journal}{\emph{Image - Music - Text}} (\bibinfo{year}{1977}), \bibinfo{pages}{32--51}.
\newblock


\bibitem[Benabdallah et~al\mbox{.}(2022)]%
        {10.1145/3532106.3533449}
\bibfield{author}{\bibinfo{person}{Gabrielle Benabdallah}, \bibinfo{person}{Ashten Alexander}, \bibinfo{person}{Sourojit Ghosh}, \bibinfo{person}{Chariell Glogovac-Smith}, \bibinfo{person}{Lacey Jacoby}, \bibinfo{person}{Caitlin Lustig}, \bibinfo{person}{Anh Nguyen}, \bibinfo{person}{Anna Parkhurst}, \bibinfo{person}{Kathryn Reyes}, \bibinfo{person}{Neilly~H. Tan}, \bibinfo{person}{Edward Wolcher}, \bibinfo{person}{Afroditi Psarra}, {and} \bibinfo{person}{Daniela Rosner}.} \bibinfo{year}{2022}\natexlab{}.
\newblock \showarticletitle{Slanted Speculations: Material Encounters with Algorithmic Bias}. In \bibinfo{booktitle}{\emph{Designing Interactive Systems Conference}} (Virtual Event, Australia) \emph{(\bibinfo{series}{DIS '22})}. \bibinfo{publisher}{Association for Computing Machinery}, \bibinfo{address}{New York, NY, USA}, \bibinfo{pages}{85–99}.
\newblock
\showISBNx{9781450393584}
\urldef\tempurl%
\url{https://doi.org/10.1145/3532106.3533449}
\showDOI{\tempurl}


\bibitem[Beretta et~al\mbox{.}(2021)]%
        {beretta2021detecting}
\bibfield{author}{\bibinfo{person}{Elena Beretta}, \bibinfo{person}{Antonio Vetr{\`o}}, \bibinfo{person}{Bruno Lepri}, {and} \bibinfo{person}{Juan Carlos~De Martin}.} \bibinfo{year}{2021}\natexlab{}.
\newblock \showarticletitle{Detecting discriminatory risk through data annotation based on bayesian inferences}. In \bibinfo{booktitle}{\emph{Proceedings of the 2021 ACM Conference on Fairness, Accountability, and Transparency}}. \bibinfo{pages}{794--804}.
\newblock


\bibitem[Bolukbasi et~al\mbox{.}(2016)]%
        {bolukbasi2016man}
\bibfield{author}{\bibinfo{person}{Tolga Bolukbasi}, \bibinfo{person}{Kai-Wei Chang}, \bibinfo{person}{James~Y Zou}, \bibinfo{person}{Venkatesh Saligrama}, {and} \bibinfo{person}{Adam~T Kalai}.} \bibinfo{year}{2016}\natexlab{}.
\newblock \showarticletitle{Man is to computer programmer as woman is to homemaker? debiasing word embeddings}.
\newblock \bibinfo{journal}{\emph{Advances in neural information processing systems}}  \bibinfo{volume}{29} (\bibinfo{year}{2016}).
\newblock


\bibitem[Bonial et~al\mbox{.}(2017)]%
        {bonial2017laying}
\bibfield{author}{\bibinfo{person}{Claire Bonial}, \bibinfo{person}{Matthew Marge}, \bibinfo{person}{Ashley Foots}, \bibinfo{person}{Felix Gervits}, \bibinfo{person}{Cory~J Hayes}, \bibinfo{person}{Cassidy Henry}, \bibinfo{person}{Susan~G Hill}, \bibinfo{person}{Anton Leuski}, \bibinfo{person}{Stephanie~M Lukin}, \bibinfo{person}{Pooja Moolchandani}, {et~al\mbox{.}}} \bibinfo{year}{2017}\natexlab{}.
\newblock \bibinfo{booktitle}{\emph{{Laying down the yellow brick road: Development of a wizard-of-oz interface for collecting human-robot dialogue}}}.
\newblock \bibinfo{publisher}{AAAI Fall Symposium Series}.
\newblock


\bibitem[Booker(2004)]%
        {booker2004seven}
\bibfield{author}{\bibinfo{person}{Christopher Booker}.} \bibinfo{year}{2004}\natexlab{}.
\newblock \bibinfo{booktitle}{\emph{The seven basic plots: Why we tell stories}}.
\newblock \bibinfo{publisher}{A\&C Black}.
\newblock


\bibitem[Braun and Clarke(2006)]%
        {braun2006using}
\bibfield{author}{\bibinfo{person}{Virginia Braun} {and} \bibinfo{person}{Victoria Clarke}.} \bibinfo{year}{2006}\natexlab{}.
\newblock \showarticletitle{Using thematic analysis in psychology}.
\newblock \bibinfo{journal}{\emph{Qualitative research in psychology}} \bibinfo{volume}{3}, \bibinfo{number}{2} (\bibinfo{year}{2006}), \bibinfo{pages}{77--101}.
\newblock


\bibitem[Buckingham(2009)]%
        {buckingham2009creative}
\bibfield{author}{\bibinfo{person}{David Buckingham}.} \bibinfo{year}{2009}\natexlab{}.
\newblock \showarticletitle{``Creative'' visual methods in media research: possibilities, problems and proposals}.
\newblock \bibinfo{journal}{\emph{Media, Culture \& Society}} \bibinfo{volume}{31}, \bibinfo{number}{4} (\bibinfo{year}{2009}), \bibinfo{pages}{633--652}.
\newblock


\bibitem[Buolamwini and Gebru(2018)]%
        {buolamwini2018gender}
\bibfield{author}{\bibinfo{person}{Joy Buolamwini} {and} \bibinfo{person}{Timnit Gebru}.} \bibinfo{year}{2018}\natexlab{}.
\newblock \showarticletitle{Gender shades: Intersectional accuracy disparities in commercial gender classification}. In \bibinfo{booktitle}{\emph{Conference on fairness, accountability and transparency}}. PMLR, \bibinfo{pages}{77--91}.
\newblock


\bibitem[Cambo and Gergle(2022)]%
        {cambo2022model}
\bibfield{author}{\bibinfo{person}{Scott~Allen Cambo} {and} \bibinfo{person}{Darren Gergle}.} \bibinfo{year}{2022}\natexlab{}.
\newblock \showarticletitle{Model Positionality and Computational Reflexivity: Promoting Reflexivity in Data Science}. In \bibinfo{booktitle}{\emph{CHI Conference on Human Factors in Computing Systems}}. \bibinfo{pages}{1--19}.
\newblock


\bibitem[Chen et~al\mbox{.}(2015)]%
        {chen2015microsoft}
\bibfield{author}{\bibinfo{person}{Xinlei Chen}, \bibinfo{person}{Hao Fang}, \bibinfo{person}{Tsung-Yi Lin}, \bibinfo{person}{Ramakrishna Vedantam}, \bibinfo{person}{Saurabh Gupta}, \bibinfo{person}{Piotr Doll{\'a}r}, {and} \bibinfo{person}{C~Lawrence Zitnick}.} \bibinfo{year}{2015}\natexlab{}.
\newblock \showarticletitle{{Microsoft COCO Captions: Data Collection and Evaluation Server}}.
\newblock \bibinfo{journal}{\emph{arXiv preprint arXiv:1504.00325}} (\bibinfo{year}{2015}).
\newblock


\bibitem[Curry and Rieser(2018)]%
        {curry2018metoo}
\bibfield{author}{\bibinfo{person}{Amanda~Cercas Curry} {and} \bibinfo{person}{Verena Rieser}.} \bibinfo{year}{2018}\natexlab{}.
\newblock \showarticletitle{\# metoo alexa: How conversational systems respond to sexual harassment}. In \bibinfo{booktitle}{\emph{Proceedings of the second acl workshop on ethics in natural language processing}}. \bibinfo{pages}{7--14}.
\newblock


\bibitem[Everingham et~al\mbox{.}(2010)]%
        {everingham2010pascal}
\bibfield{author}{\bibinfo{person}{Mark Everingham}, \bibinfo{person}{Luc Van~Gool}, \bibinfo{person}{Christopher~KI Williams}, \bibinfo{person}{John Winn}, {and} \bibinfo{person}{Andrew Zisserman}.} \bibinfo{year}{2010}\natexlab{}.
\newblock \showarticletitle{The pascal visual object classes (voc) challenge}.
\newblock \bibinfo{journal}{\emph{International journal of computer vision}} \bibinfo{volume}{88}, \bibinfo{number}{2} (\bibinfo{year}{2010}), \bibinfo{pages}{303--338}.
\newblock


\bibitem[Ferraro et~al\mbox{.}(2015)]%
        {ferraro2015survey}
\bibfield{author}{\bibinfo{person}{Francis Ferraro}, \bibinfo{person}{Nasrin Mostafazadeh}, \bibinfo{person}{Lucy Vanderwende}, \bibinfo{person}{Jacob Devlin}, \bibinfo{person}{Michel Galley}, \bibinfo{person}{Margaret Mitchell}, {et~al\mbox{.}}} \bibinfo{year}{2015}\natexlab{}.
\newblock \showarticletitle{{A Survey of Current Datasets for Vision and Language Research}}.
\newblock \bibinfo{journal}{\emph{arXiv preprint arXiv:1506.06833}} (\bibinfo{year}{2015}).
\newblock


\bibitem[Geiger et~al\mbox{.}(2021)]%
        {geiger2021garbage}
\bibfield{author}{\bibinfo{person}{R~Stuart Geiger}, \bibinfo{person}{Dominique Cope}, \bibinfo{person}{Jamie Ip}, \bibinfo{person}{Marsha Lotosh}, \bibinfo{person}{Aayush Shah}, \bibinfo{person}{Jenny Weng}, {and} \bibinfo{person}{Rebekah Tang}.} \bibinfo{year}{2021}\natexlab{}.
\newblock \showarticletitle{“Garbage in, garbage out” revisited: What do machine learning application papers report about human-labeled training data?}
\newblock \bibinfo{journal}{\emph{Quantitative Science Studies}} \bibinfo{volume}{2}, \bibinfo{number}{3} (\bibinfo{year}{2021}), \bibinfo{pages}{795--827}.
\newblock


\bibitem[Gordon and Spierling(2018)]%
        {gordon2018playing}
\bibfield{author}{\bibinfo{person}{Andrew~S Gordon} {and} \bibinfo{person}{Ulrike Spierling}.} \bibinfo{year}{2018}\natexlab{}.
\newblock \showarticletitle{Playing story creation games with logical abduction}. In \bibinfo{booktitle}{\emph{International Conference on Interactive Digital Storytelling}}. Springer, \bibinfo{pages}{478--482}.
\newblock


\bibitem[Halperin(2022)]%
        {10.1145/3529705}
\bibfield{author}{\bibinfo{person}{Brett~A. Halperin}.} \bibinfo{year}{2022}\natexlab{}.
\newblock \showarticletitle{Airbrush Hyperfabric: Designing Interactive Storytelling Fabric Connected to Motion Graphics and Music}.
\newblock \bibinfo{journal}{\emph{Interactions}} \bibinfo{volume}{29}, \bibinfo{number}{3} (\bibinfo{date}{apr} \bibinfo{year}{2022}), \bibinfo{pages}{8–9}.
\newblock
\showISSN{1072-5520}
\urldef\tempurl%
\url{https://doi.org/10.1145/3529705}
\showDOI{\tempurl}


\bibitem[Halperin et~al\mbox{.}(2023)]%
        {Halperin2023}
\bibfield{author}{\bibinfo{person}{Brett~A. Halperin}, \bibinfo{person}{Gary Hsieh}, \bibinfo{person}{Erin McElroy}, \bibinfo{person}{James Pierce}, {and} \bibinfo{person}{Daniela~K. Rosner}.} \bibinfo{year}{2023}\natexlab{}.
\newblock \showarticletitle{Probing a Community-Based Conversational Storytelling Agent to Document Digital Stories of Housing Insecurity}. In \bibinfo{booktitle}{\emph{Proceedings of the 2023 CHI Conference on Human Factors in Computing Systems}} (Hamburg, Germany) \emph{(\bibinfo{series}{CHI '23})}. \bibinfo{publisher}{Association for Computing Machinery}, \bibinfo{address}{New York, NY, USA}.
\newblock
\urldef\tempurl%
\url{https://doi.org/10.1145/3544548.3581109}
\showDOI{\tempurl}


\bibitem[Hartsook et~al\mbox{.}(2011)]%
        {hartsook2011toward}
\bibfield{author}{\bibinfo{person}{Ken Hartsook}, \bibinfo{person}{Alexander Zook}, \bibinfo{person}{Sauvik Das}, {and} \bibinfo{person}{Mark~O Riedl}.} \bibinfo{year}{2011}\natexlab{}.
\newblock \showarticletitle{Toward supporting stories with procedurally generated game worlds}. In \bibinfo{booktitle}{\emph{2011 IEEE Conference on Computational Intelligence and Games (CIG'11)}}. IEEE, \bibinfo{pages}{297--304}.
\newblock


\bibitem[Hermanto(2019)]%
        {hermanto2019visual}
\bibfield{author}{\bibinfo{person}{Yon Ade~Lose Hermanto}.} \bibinfo{year}{2019}\natexlab{}.
\newblock \showarticletitle{Visual storytelling in folklore children book illustration}.
\newblock \bibinfo{journal}{\emph{Asian Journal of Research in Education and Social Sciences}} \bibinfo{volume}{1}, \bibinfo{number}{1} (\bibinfo{year}{2019}), \bibinfo{pages}{62--70}.
\newblock


\bibitem[Hill et~al\mbox{.}(2015)]%
        {hill2015goldilocks}
\bibfield{author}{\bibinfo{person}{Felix Hill}, \bibinfo{person}{Antoine Bordes}, \bibinfo{person}{Sumit Chopra}, {and} \bibinfo{person}{Jason Weston}.} \bibinfo{year}{2015}\natexlab{}.
\newblock \showarticletitle{The goldilocks principle: Reading children's books with explicit memory representations}.
\newblock \bibinfo{journal}{\emph{arXiv preprint arXiv:1511.02301}} (\bibinfo{year}{2015}).
\newblock


\bibitem[Hodosh et~al\mbox{.}(2013)]%
        {hodosh2013framing}
\bibfield{author}{\bibinfo{person}{Micah Hodosh}, \bibinfo{person}{Peter Young}, {and} \bibinfo{person}{Julia Hockenmaier}.} \bibinfo{year}{2013}\natexlab{}.
\newblock \showarticletitle{{Framing Image Description as a Ranking Task: Data, Models and Evaluation Metrics}}.
\newblock \bibinfo{journal}{\emph{Journal of Artificial Intelligence Research}}  \bibinfo{volume}{47} (\bibinfo{year}{2013}), \bibinfo{pages}{853--899}.
\newblock


\bibitem[Holtzman et~al\mbox{.}(2020)]%
        {holtzman2019curious}
\bibfield{author}{\bibinfo{person}{Ari Holtzman}, \bibinfo{person}{Jan Buys}, \bibinfo{person}{Li Du}, \bibinfo{person}{Maxwell Forbes}, {and} \bibinfo{person}{Yejin Choi}.} \bibinfo{year}{2020}\natexlab{}.
\newblock \showarticletitle{The Curious Case of Neural Text Degeneration}. In \bibinfo{booktitle}{\emph{International Conference on Learning Representations}}.
\newblock


\bibitem[Huang et~al\mbox{.}(2016)]%
        {huang2016visual}
\bibfield{author}{\bibinfo{person}{Ting-Hao Huang}, \bibinfo{person}{Francis Ferraro}, \bibinfo{person}{Nasrin Mostafazadeh}, \bibinfo{person}{Ishan Misra}, \bibinfo{person}{Aishwarya Agrawal}, \bibinfo{person}{Jacob Devlin}, \bibinfo{person}{Ross Girshick}, \bibinfo{person}{Xiaodong He}, \bibinfo{person}{Pushmeet Kohli}, \bibinfo{person}{Dhruv Batra}, {et~al\mbox{.}}} \bibinfo{year}{2016}\natexlab{}.
\newblock \showarticletitle{Visual storytelling}. In \bibinfo{booktitle}{\emph{Conference of the North American Chapter of the Association for Computational Linguistics: Human Language Technologies}}. \bibinfo{pages}{1233--1239}.
\newblock


\bibitem[Jana(2019)]%
        {pix2story}
\bibfield{author}{\bibinfo{person}{Tara Jana}.} \bibinfo{year}{2019}\natexlab{}.
\newblock \bibinfo{title}{Pix2Story: Neural storyteller which creates machine-generated story in several literature genre}.
\newblock \bibinfo{howpublished}{\url{https://azure.microsoft.com/en-us/blog/pix2story-neural-storyteller-which-creates-machine-generated-story-in-several-literature-genre/}}.
\newblock


\bibitem[Karimi et~al\mbox{.}(2019)]%
        {10.1145/3325480.3325488}
\bibfield{author}{\bibinfo{person}{Pegah Karimi}, \bibinfo{person}{Nicholas Davis}, \bibinfo{person}{Mary~Lou Maher}, \bibinfo{person}{Kazjon Grace}, {and} \bibinfo{person}{Lina Lee}.} \bibinfo{year}{2019}\natexlab{}.
\newblock \showarticletitle{Relating Cognitive Models of Design Creativity to the Similarity of Sketches Generated by an AI Partner}. In \bibinfo{booktitle}{\emph{Creativity and Cognition}} (San Diego, CA, USA) \emph{(\bibinfo{series}{C\&C '19})}. \bibinfo{publisher}{Association for Computing Machinery}, \bibinfo{address}{New York, NY, USA}, \bibinfo{pages}{259–270}.
\newblock
\showISBNx{9781450359177}
\urldef\tempurl%
\url{https://doi.org/10.1145/3325480.3325488}
\showDOI{\tempurl}


\bibitem[Karimi et~al\mbox{.}(2018)]%
        {karimi2018creative}
\bibfield{author}{\bibinfo{person}{Pegah Karimi}, \bibinfo{person}{Kazjon Grace}, \bibinfo{person}{Nicholas Davis}, {and} \bibinfo{person}{Mary~Lou Maher}.} \bibinfo{year}{2018}\natexlab{}.
\newblock \showarticletitle{Creative sketching apprentice: Supporting conceptual shifts in sketch ideation}. In \bibinfo{booktitle}{\emph{International Conference on-Design Computing and Cognition}}. Springer, \bibinfo{pages}{721--738}.
\newblock


\bibitem[Kerwin et~al\mbox{.}(2020)]%
        {kerwin2020us}
\bibfield{author}{\bibinfo{person}{Donald Kerwin}, \bibinfo{person}{Mike Nicholson}, \bibinfo{person}{Daniela Alulema}, {and} \bibinfo{person}{Robert Warren}.} \bibinfo{year}{2020}\natexlab{}.
\newblock \showarticletitle{US foreign-born essential workers by status and state, and the global pandemic}.
\newblock \bibinfo{journal}{\emph{Center for Migration Studies}} (\bibinfo{year}{2020}).
\newblock


\bibitem[Kiritchenko and Mohammad(2018)]%
        {kiritchenko2018examining}
\bibfield{author}{\bibinfo{person}{Svetlana Kiritchenko} {and} \bibinfo{person}{Saif Mohammad}.} \bibinfo{year}{2018}\natexlab{}.
\newblock \showarticletitle{Examining Gender and Race Bias in Two Hundred Sentiment Analysis Systems}. In \bibinfo{booktitle}{\emph{Proceedings of the Seventh Joint Conference on Lexical and Computational Semantics}}. \bibinfo{pages}{43--53}.
\newblock


\bibitem[Kiros et~al\mbox{.}(2015)]%
        {kiros2015skip}
\bibfield{author}{\bibinfo{person}{Ryan Kiros}, \bibinfo{person}{Yukun Zhu}, \bibinfo{person}{Ruslan Salakhutdinov}, \bibinfo{person}{Richard~S Zemel}, \bibinfo{person}{Antonio Torralba}, \bibinfo{person}{Raquel Urtasun}, {and} \bibinfo{person}{Sanja Fidler}.} \bibinfo{year}{2015}\natexlab{}.
\newblock \bibinfo{booktitle}{\emph{{Skip-Thought vectors}}}.
\newblock \bibinfo{publisher}{arXiv:1506.06726}.
\newblock


\bibitem[Koller(2005)]%
        {koller2005michael}
\bibfield{author}{\bibinfo{person}{Veronika Koller}.} \bibinfo{year}{2005}\natexlab{}.
\newblock \showarticletitle{Michael Mateas and Phoebe Sengers (eds). 2003. Narrative Intelligence}.
\newblock \bibinfo{journal}{\emph{Studies in Language. International Journal sponsored by the Foundation “Foundations of Language”}} \bibinfo{volume}{29}, \bibinfo{number}{1} (\bibinfo{year}{2005}), \bibinfo{pages}{227--234}.
\newblock


\bibitem[Korsgaard et~al\mbox{.}(2022)]%
        {Korsgaard2022}
\bibfield{author}{\bibinfo{person}{Henrik Korsgaard}, \bibinfo{person}{Peter Lyle}, \bibinfo{person}{Joanna Saad-Sulonen}, \bibinfo{person}{Clemens~Nylandsted Klokmose}, \bibinfo{person}{Midas Nouwens}, {and} \bibinfo{person}{Susanne B\o{}dker}.} \bibinfo{year}{2022}\natexlab{}.
\newblock \showarticletitle{Collectives and Their Artifact Ecologies}.
\newblock \bibinfo{journal}{\emph{Proc. ACM Hum.-Comput. Interact.}} \bibinfo{volume}{6}, \bibinfo{number}{CSCW2}, Article \bibinfo{articleno}{432} (\bibinfo{date}{nov} \bibinfo{year}{2022}), \bibinfo{numpages}{26}~pages.
\newblock
\urldef\tempurl%
\url{https://doi.org/10.1145/3555533}
\showDOI{\tempurl}


\bibitem[Kreminski et~al\mbox{.}(2020)]%
        {10.1145/3402942.3402953}
\bibfield{author}{\bibinfo{person}{Max Kreminski}, \bibinfo{person}{Melanie Dickinson}, \bibinfo{person}{Michael Mateas}, {and} \bibinfo{person}{Noah Wardrip-Fruin}.} \bibinfo{year}{2020}\natexlab{}.
\newblock \showarticletitle{Why Are We Like This?: The AI Architecture of a Co-Creative Storytelling Game}. In \bibinfo{booktitle}{\emph{International Conference on the Foundations of Digital Games}} (Bugibba, Malta) \emph{(\bibinfo{series}{FDG '20})}. \bibinfo{publisher}{Association for Computing Machinery}, \bibinfo{address}{New York, NY, USA}, Article \bibinfo{articleno}{13}, \bibinfo{numpages}{4}~pages.
\newblock
\showISBNx{9781450388078}
\urldef\tempurl%
\url{https://doi.org/10.1145/3402942.3402953}
\showDOI{\tempurl}


\bibitem[Kreminski and Wardrip-Fruin(2019)]%
        {10.1145/3337722.3341861}
\bibfield{author}{\bibinfo{person}{Max Kreminski} {and} \bibinfo{person}{Noah Wardrip-Fruin}.} \bibinfo{year}{2019}\natexlab{}.
\newblock \showarticletitle{Generative Games as Storytelling Partners}. In \bibinfo{booktitle}{\emph{International Conference on the Foundations of Digital Games}} (San Luis Obispo, California, USA) \emph{(\bibinfo{series}{FDG '19})}. \bibinfo{publisher}{Association for Computing Machinery}, \bibinfo{address}{New York, NY, USA}, Article \bibinfo{articleno}{103}, \bibinfo{numpages}{8}~pages.
\newblock
\showISBNx{9781450372176}
\urldef\tempurl%
\url{https://doi.org/10.1145/3337722.3341861}
\showDOI{\tempurl}


\bibitem[Krishna et~al\mbox{.}(2017)]%
        {krishna2017visual}
\bibfield{author}{\bibinfo{person}{Ranjay Krishna}, \bibinfo{person}{Yuke Zhu}, \bibinfo{person}{Oliver Groth}, \bibinfo{person}{Justin Johnson}, \bibinfo{person}{Kenji Hata}, \bibinfo{person}{Joshua Kravitz}, \bibinfo{person}{Stephanie Chen}, \bibinfo{person}{Yannis Kalantidis}, \bibinfo{person}{Li-Jia Li}, \bibinfo{person}{David~A Shamma}, {et~al\mbox{.}}} \bibinfo{year}{2017}\natexlab{}.
\newblock \showarticletitle{Visual genome: Connecting language and vision using crowdsourced dense image annotations}.
\newblock \bibinfo{journal}{\emph{International journal of computer vision}} \bibinfo{volume}{123}, \bibinfo{number}{1} (\bibinfo{year}{2017}), \bibinfo{pages}{32--73}.
\newblock


\bibitem[Lee(2016)]%
        {lee2016tay}
\bibfield{author}{\bibinfo{person}{Peter Lee}.} \bibinfo{year}{2016}\natexlab{}.
\newblock \bibinfo{booktitle}{\emph{Learning from Tay’s introduction}}.
\newblock
\urldef\tempurl%
\url{https://blogs.microsoft.com/blog/2016/03/25/learning-tays-introduction/}
\showURL{%
\tempurl}


\bibitem[Li et~al\mbox{.}(2022)]%
        {li2022clip}
\bibfield{author}{\bibinfo{person}{Manling Li}, \bibinfo{person}{Ruochen Xu}, \bibinfo{person}{Shuohang Wang}, \bibinfo{person}{Luowei Zhou}, \bibinfo{person}{Xudong Lin}, \bibinfo{person}{Chenguang Zhu}, \bibinfo{person}{Michael Zeng}, \bibinfo{person}{Heng Ji}, {and} \bibinfo{person}{Shih-Fu Chang}.} \bibinfo{year}{2022}\natexlab{}.
\newblock \showarticletitle{Clip-event: Connecting text and images with event structures}. In \bibinfo{booktitle}{\emph{Proceedings of the IEEE/CVF Conference on Computer Vision and Pattern Recognition}}. \bibinfo{pages}{16420--16429}.
\newblock


\bibitem[Li et~al\mbox{.}(2020)]%
        {li2020gaia}
\bibfield{author}{\bibinfo{person}{Manling Li}, \bibinfo{person}{Alireza Zareian}, \bibinfo{person}{Ying Lin}, \bibinfo{person}{Xiaoman Pan}, \bibinfo{person}{Spencer Whitehead}, \bibinfo{person}{Brian Chen}, \bibinfo{person}{Bo Wu}, \bibinfo{person}{Heng Ji}, \bibinfo{person}{Shih-Fu Chang}, \bibinfo{person}{Clare Voss}, {et~al\mbox{.}}} \bibinfo{year}{2020}\natexlab{}.
\newblock \showarticletitle{Gaia: A fine-grained multimedia knowledge extraction system}. In \bibinfo{booktitle}{\emph{Proceedings of the 58th Annual Meeting of the Association for Computational Linguistics: System Demonstrations}}. \bibinfo{pages}{77--86}.
\newblock


\bibitem[Lin et~al\mbox{.}(2014)]%
        {lin2014microsoft}
\bibfield{author}{\bibinfo{person}{Tsung-Yi Lin}, \bibinfo{person}{Michael Maire}, \bibinfo{person}{Serge Belongie}, \bibinfo{person}{James Hays}, \bibinfo{person}{Pietro Perona}, \bibinfo{person}{Deva Ramanan}, \bibinfo{person}{Piotr Doll{\'a}r}, {and} \bibinfo{person}{C~Lawrence Zitnick}.} \bibinfo{year}{2014}\natexlab{}.
\newblock \showarticletitle{Microsoft coco: Common objects in context}. In \bibinfo{booktitle}{\emph{European conference on computer vision}}. Springer, \bibinfo{pages}{740--755}.
\newblock


\bibitem[Lin et~al\mbox{.}(2020)]%
        {10.1145/3313831.3376258}
\bibfield{author}{\bibinfo{person}{Yuyu Lin}, \bibinfo{person}{Jiahao Guo}, \bibinfo{person}{Yang Chen}, \bibinfo{person}{Cheng Yao}, {and} \bibinfo{person}{Fangtian Ying}.} \bibinfo{year}{2020}\natexlab{}.
\newblock \showarticletitle{It Is Your Turn: Collaborative Ideation With a Co-Creative Robot through Sketch}. In \bibinfo{booktitle}{\emph{Proceedings of the 2020 CHI Conference on Human Factors in Computing Systems}} (Honolulu, HI, USA) \emph{(\bibinfo{series}{CHI '20})}. \bibinfo{publisher}{Association for Computing Machinery}, \bibinfo{address}{New York, NY, USA}, \bibinfo{pages}{1–14}.
\newblock
\showISBNx{9781450367080}
\urldef\tempurl%
\url{https://doi.org/10.1145/3313831.3376258}
\showDOI{\tempurl}


\bibitem[Lukin et~al\mbox{.}(2018)]%
        {lukin2018pipeline}
\bibfield{author}{\bibinfo{person}{Stephanie Lukin}, \bibinfo{person}{Reginald Hobbs}, {and} \bibinfo{person}{Clare Voss}.} \bibinfo{year}{2018}\natexlab{}.
\newblock \showarticletitle{A Pipeline for Creative Visual Storytelling}. In \bibinfo{booktitle}{\emph{Proceedings of the First Workshop on Storytelling}}. \bibinfo{pages}{20--32}.
\newblock


\bibitem[Lukin and Eum(2023)]%
        {lukin2022controllable}
\bibfield{author}{\bibinfo{person}{Stephanie~M. Lukin} {and} \bibinfo{person}{Sungmin Eum}.} \bibinfo{year}{2023}\natexlab{}.
\newblock \showarticletitle{{Controllable Narrative Generation from Images}}. In \bibinfo{booktitle}{\emph{AAAI Creative AI Across Modalities Workshop}}. Springer.
\newblock


\bibitem[Madden(99)]%
        {madden99ways}
\bibfield{author}{\bibinfo{person}{Matt Madden}.} \bibinfo{year}{99}\natexlab{}.
\newblock \showarticletitle{ways to tell a story: Exercises in style}.
\newblock \bibinfo{journal}{\emph{New York, NY: Chamberlain Bros}} (\bibinfo{year}{99}).
\newblock


\bibitem[Marge et~al\mbox{.}(2016)]%
        {marge2016applying}
\bibfield{author}{\bibinfo{person}{Matthew Marge}, \bibinfo{person}{Claire Bonial}, \bibinfo{person}{Brendan Byrne}, \bibinfo{person}{Taylor Cassidy}, \bibinfo{person}{A~William Evans}, \bibinfo{person}{Susan~G Hill}, {and} \bibinfo{person}{Clare Voss}.} \bibinfo{year}{2016}\natexlab{}.
\newblock \showarticletitle{Applying the Wizard-Of-Oz Technique to Multimodal Human-Robot Dialogue}. In \bibinfo{booktitle}{\emph{Proceedings of the IEEE International Symposium on Robot and Human Interactive Communication (RO-MAN)}}.
\newblock


\bibitem[Mateas and Stern(2003)]%
        {mateas2003integrating}
\bibfield{author}{\bibinfo{person}{Michael Mateas} {and} \bibinfo{person}{Andrew Stern}.} \bibinfo{year}{2003}\natexlab{}.
\newblock \showarticletitle{Integrating plot, character and natural language processing in the interactive drama Fa{\c{c}}ade}. In \bibinfo{booktitle}{\emph{Proceedings of the 1st International Conference on Technologies for Interactive Digital Storytelling and Entertainment (TIDSE-03)}}, Vol.~\bibinfo{volume}{2}.
\newblock


\bibitem[McCloud(1993)]%
        {mccloud1993understanding}
\bibfield{author}{\bibinfo{person}{Scott McCloud}.} \bibinfo{year}{1993}\natexlab{}.
\newblock \showarticletitle{Understanding comics: The invisible art}.
\newblock \bibinfo{journal}{\emph{Northampton, Mass}} (\bibinfo{year}{1993}).
\newblock


\bibitem[McCloud and Manning(1998)]%
        {mccloud1998understanding}
\bibfield{author}{\bibinfo{person}{Scott McCloud} {and} \bibinfo{person}{AD Manning}.} \bibinfo{year}{1998}\natexlab{}.
\newblock \showarticletitle{Understanding comics: The invisible art}.
\newblock \bibinfo{journal}{\emph{IEEE Transactions on Professional Communications}} \bibinfo{volume}{41}, \bibinfo{number}{1} (\bibinfo{year}{1998}), \bibinfo{pages}{66--69}.
\newblock


\bibitem[Miles and Huberman(1994)]%
        {miles1994qualitative}
\bibfield{author}{\bibinfo{person}{Matthew~B Miles} {and} \bibinfo{person}{A~Michael Huberman}.} \bibinfo{year}{1994}\natexlab{}.
\newblock \bibinfo{booktitle}{\emph{Qualitative data analysis: An expanded sourcebook}}.
\newblock \bibinfo{publisher}{sage}.
\newblock


\bibitem[Mostafazadeh et~al\mbox{.}(2016)]%
        {mostafazadeh2016corpus}
\bibfield{author}{\bibinfo{person}{Nasrin Mostafazadeh}, \bibinfo{person}{Nathanael Chambers}, \bibinfo{person}{Xiaodong He}, \bibinfo{person}{Devi Parikh}, \bibinfo{person}{Dhruv Batra}, \bibinfo{person}{Lucy Vanderwende}, \bibinfo{person}{Pushmeet Kohli}, {and} \bibinfo{person}{James Allen}.} \bibinfo{year}{2016}\natexlab{}.
\newblock \showarticletitle{A corpus and cloze evaluation for deeper understanding of commonsense stories}. In \bibinfo{booktitle}{\emph{Proceedings of the 2016 Conference of the North American Chapter of the Association for Computational Linguistics: Human Language Technologies}}. \bibinfo{pages}{839--849}.
\newblock


\bibitem[Murray(2011)]%
        {murray2011inventing}
\bibfield{author}{\bibinfo{person}{Janet~H Murray}.} \bibinfo{year}{2011}\natexlab{}.
\newblock \bibinfo{booktitle}{\emph{Inventing the medium: principles of interaction design as a cultural practice}}.
\newblock \bibinfo{publisher}{Mit Press}.
\newblock


\bibitem[Murray(2017)]%
        {murray2017hamlet}
\bibfield{author}{\bibinfo{person}{Janet~H Murray}.} \bibinfo{year}{2017}\natexlab{}.
\newblock \bibinfo{booktitle}{\emph{Hamlet on the Holodeck, updated edition: The Future of Narrative in Cyberspace}}.
\newblock \bibinfo{publisher}{MIT press}.
\newblock


\bibitem[Noble(2018)]%
        {noble2018algorithms}
\bibfield{author}{\bibinfo{person}{Safiya~Umoja Noble}.} \bibinfo{year}{2018}\natexlab{}.
\newblock \showarticletitle{Algorithms of oppression}.
\newblock In \bibinfo{booktitle}{\emph{Algorithms of Oppression}}. \bibinfo{publisher}{New York University Press}.
\newblock


\bibitem[O'Neill and Riedl(2014)]%
        {o2014applying}
\bibfield{author}{\bibinfo{person}{Brian O'Neill} {and} \bibinfo{person}{Mark Riedl}.} \bibinfo{year}{2014}\natexlab{}.
\newblock \showarticletitle{Applying qualitative research methods to narrative knowledge engineering}. In \bibinfo{booktitle}{\emph{2014 Workshop on Computational Models of Narrative}}. Schloss Dagstuhl-Leibniz-Zentrum fuer Informatik.
\newblock


\bibitem[OpenAI(2022)]%
        {chatgpt}
\bibfield{author}{\bibinfo{person}{OpenAI}.} \bibinfo{year}{2022}\natexlab{}.
\newblock \bibinfo{title}{ChatGPT: Optimizing Language Models for Dialogue}.
\newblock \bibinfo{howpublished}{\url{https://openai.com/blog/chatgpt/}}.
\newblock


\bibitem[Ordonez et~al\mbox{.}(2011)]%
        {ordonez2011im2text}
\bibfield{author}{\bibinfo{person}{Vicente Ordonez}, \bibinfo{person}{Girish Kulkarni}, {and} \bibinfo{person}{Tamara Berg}.} \bibinfo{year}{2011}\natexlab{}.
\newblock \showarticletitle{Im2text: Describing images using 1 million captioned photographs}.
\newblock \bibinfo{journal}{\emph{Advances in neural information processing systems}}  \bibinfo{volume}{24} (\bibinfo{year}{2011}).
\newblock


\bibitem[Park and Kim(2015)]%
        {park2015expressing}
\bibfield{author}{\bibinfo{person}{Cesc~C Park} {and} \bibinfo{person}{Gunhee Kim}.} \bibinfo{year}{2015}\natexlab{}.
\newblock \showarticletitle{Expressing an image stream with a sequence of natural sentences}.
\newblock \bibinfo{journal}{\emph{Advances in neural information processing systems}}  \bibinfo{volume}{28} (\bibinfo{year}{2015}).
\newblock


\bibitem[Plummer et~al\mbox{.}(2015)]%
        {plummer2015flickr30k}
\bibfield{author}{\bibinfo{person}{Bryan~A Plummer}, \bibinfo{person}{Liwei Wang}, \bibinfo{person}{Chris~M Cervantes}, \bibinfo{person}{Juan~C Caicedo}, \bibinfo{person}{Julia Hockenmaier}, {and} \bibinfo{person}{Svetlana Lazebnik}.} \bibinfo{year}{2015}\natexlab{}.
\newblock \showarticletitle{Flickr30k entities: Collecting region-to-phrase correspondences for richer image-to-sentence models}. In \bibinfo{booktitle}{\emph{Proceedings of the IEEE international conference on computer vision}}. \bibinfo{pages}{2641--2649}.
\newblock


\bibitem[Rashtchian et~al\mbox{.}(2010)]%
        {rashtchian2010collecting}
\bibfield{author}{\bibinfo{person}{Cyrus Rashtchian}, \bibinfo{person}{Peter Young}, \bibinfo{person}{Micah Hodosh}, {and} \bibinfo{person}{Julia Hockenmaier}.} \bibinfo{year}{2010}\natexlab{}.
\newblock \showarticletitle{{Collecting Image Annotations using Amazon's Mechanical Turk}}. In \bibinfo{booktitle}{\emph{Proceedings of the NAACL HLT 2010 Workshop on Creating Speech and Language Data with Amazon's Mechanical Turk}}. Association for Computational Linguistics, \bibinfo{pages}{139--147}.
\newblock


\bibitem[Roemmele et~al\mbox{.}(2011)]%
        {roemmele2011choice}
\bibfield{author}{\bibinfo{person}{Melissa Roemmele}, \bibinfo{person}{Cosmin~Adrian Bejan}, {and} \bibinfo{person}{Andrew~S Gordon}.} \bibinfo{year}{2011}\natexlab{}.
\newblock \showarticletitle{Choice of Plausible Alternatives: An Evaluation of Commonsense Causal Reasoning.}. In \bibinfo{booktitle}{\emph{AAAI spring symposium: logical formalizations of commonsense reasoning}}. \bibinfo{pages}{90--95}.
\newblock


\bibitem[Rosner(2018)]%
        {rosner2018critical}
\bibfield{author}{\bibinfo{person}{Daniela~K Rosner}.} \bibinfo{year}{2018}\natexlab{}.
\newblock \bibinfo{booktitle}{\emph{Critical fabulations: Reworking the methods and margins of design}}.
\newblock \bibinfo{publisher}{MIT Press}.
\newblock


\bibitem[Rosner et~al\mbox{.}(2018)]%
        {10.1145/3173574.3174105}
\bibfield{author}{\bibinfo{person}{Daniela~K. Rosner}, \bibinfo{person}{Samantha Shorey}, \bibinfo{person}{Brock~R. Craft}, {and} \bibinfo{person}{Helen Remick}.} \bibinfo{year}{2018}\natexlab{}.
\newblock \showarticletitle{Making Core Memory: Design Inquiry into Gendered Legacies of Engineering and Craftwork}. In \bibinfo{booktitle}{\emph{Proceedings of the 2018 CHI Conference on Human Factors in Computing Systems}} (Montreal QC, Canada) \emph{(\bibinfo{series}{CHI '18})}. \bibinfo{publisher}{Association for Computing Machinery}, \bibinfo{address}{New York, NY, USA}, \bibinfo{pages}{1–13}.
\newblock
\showISBNx{9781450356206}
\urldef\tempurl%
\url{https://doi.org/10.1145/3173574.3174105}
\showDOI{\tempurl}


\bibitem[Singh et~al\mbox{.}(2018)]%
        {kumar2018dock}
\bibfield{author}{\bibinfo{person}{Kumar Singh}, \bibinfo{person}{Krishna}, \bibinfo{person}{Santosh Divvala}, \bibinfo{person}{Ali Farhadi}, {and} \bibinfo{person}{Yong Jae~Lee}.} \bibinfo{year}{2018}\natexlab{}.
\newblock \showarticletitle{{DOCK: Detecting Objects by transferring Common-sense Knowledge}}. In \bibinfo{booktitle}{\emph{Proceedings of the European Conference on Computer Vision (ECCV)}}. \bibinfo{pages}{492--508}.
\newblock


\bibitem[Smiley et~al\mbox{.}(2017)]%
        {smiley2017say}
\bibfield{author}{\bibinfo{person}{Charese Smiley}, \bibinfo{person}{Frank Schilder}, \bibinfo{person}{Vassilis Plachouras}, {and} \bibinfo{person}{Jochen~L Leidner}.} \bibinfo{year}{2017}\natexlab{}.
\newblock \showarticletitle{Say the right thing right: Ethics issues in natural language generation systems}. In \bibinfo{booktitle}{\emph{ACL Workshop on Ethics in Natural Language Processing}}. \bibinfo{pages}{103--108}.
\newblock


\bibitem[Thorne(1987)]%
        {thorne1987press}
\bibfield{author}{\bibinfo{person}{Avril Thorne}.} \bibinfo{year}{1987}\natexlab{}.
\newblock \showarticletitle{The press of personality: A study of conversations between introverts and extraverts.}
\newblock \bibinfo{journal}{\emph{Journal of Personality and Social Psychology}} \bibinfo{volume}{53}, \bibinfo{number}{4} (\bibinfo{year}{1987}), \bibinfo{pages}{718}.
\newblock


\bibitem[Thorne and McLean(2003)]%
        {thorne2003telling}
\bibfield{author}{\bibinfo{person}{Avril Thorne} {and} \bibinfo{person}{Kate~C McLean}.} \bibinfo{year}{2003}\natexlab{}.
\newblock \showarticletitle{Telling traumatic events in adolescence: A study of master narrative positioning}.
\newblock \bibinfo{journal}{\emph{Connecting culture and memory: The development of an autobiographical self}} (\bibinfo{year}{2003}), \bibinfo{pages}{169--185}.
\newblock


\bibitem[Valls-Vargas(2013)]%
        {valls2013narrative}
\bibfield{author}{\bibinfo{person}{Josep Valls-Vargas}.} \bibinfo{year}{2013}\natexlab{}.
\newblock \showarticletitle{Narrative extraction, processing and generation for interactive fiction and computer games}. In \bibinfo{booktitle}{\emph{Ninth Artificial Intelligence and Interactive Digital Entertainment Conference}}.
\newblock


\bibitem[van Van~Looy and Baetens(2003)]%
        {van2003close}
\bibfield{author}{\bibinfo{person}{Jan van Van~Looy} {and} \bibinfo{person}{Jan Baetens}.} \bibinfo{year}{2003}\natexlab{}.
\newblock \bibinfo{booktitle}{\emph{Close reading new media: Analyzing electronic literature}}. Vol.~\bibinfo{volume}{16}.
\newblock \bibinfo{publisher}{Leuven University Press}.
\newblock


\bibitem[Vinyals et~al\mbox{.}(2015)]%
        {vinyals2015show}
\bibfield{author}{\bibinfo{person}{Oriol Vinyals}, \bibinfo{person}{Alexander Toshev}, \bibinfo{person}{Samy Bengio}, {and} \bibinfo{person}{Dumitru Erhan}.} \bibinfo{year}{2015}\natexlab{}.
\newblock \showarticletitle{Show and tell: A neural image caption generator}. In \bibinfo{booktitle}{\emph{Proceedings of the IEEE conference on computer vision and pattern recognition}}. \bibinfo{pages}{3156--3164}.
\newblock


\bibitem[Wang et~al\mbox{.}(2018)]%
        {wang2018no}
\bibfield{author}{\bibinfo{person}{Xin Wang}, \bibinfo{person}{Wenhu Chen}, \bibinfo{person}{Yuan-Fang Wang}, {and} \bibinfo{person}{William~Yang Wang}.} \bibinfo{year}{2018}\natexlab{}.
\newblock \showarticletitle{No Metrics Are Perfect: Adversarial Reward Learning for Visual Storytelling}. In \bibinfo{booktitle}{\emph{Proc. of the Association for Computational Linguistics}}. \bibinfo{pages}{899--909}.
\newblock


\bibitem[Wardrip-Fruin(2009)]%
        {wardrip2009expressive}
\bibfield{author}{\bibinfo{person}{Noah Wardrip-Fruin}.} \bibinfo{year}{2009}\natexlab{}.
\newblock \bibinfo{booktitle}{\emph{Expressive Processing: Digital fictions, computer games, and software studies}}.
\newblock \bibinfo{publisher}{MIT press}.
\newblock


\bibitem[Wolf et~al\mbox{.}(2017)]%
        {wolf2017we}
\bibfield{author}{\bibinfo{person}{Marty~J Wolf}, \bibinfo{person}{Keith~W Miller}, {and} \bibinfo{person}{Frances~S Grodzinsky}.} \bibinfo{year}{2017}\natexlab{}.
\newblock \showarticletitle{Why we should have seen that coming: comments on microsoft’s tay “experiment,” and wider implications}.
\newblock \bibinfo{journal}{\emph{The ORBIT Journal}} \bibinfo{volume}{1}, \bibinfo{number}{2} (\bibinfo{year}{2017}), \bibinfo{pages}{1--12}.
\newblock


\bibitem[Wright and Queneau(1986)]%
        {wright1986exercises}
\bibfield{author}{\bibinfo{person}{Barbara Wright} {and} \bibinfo{person}{R Queneau}.} \bibinfo{year}{1986}\natexlab{}.
\newblock \bibinfo{title}{Exercises in style}.
\newblock
\newblock


\bibitem[Yatskar et~al\mbox{.}(2016)]%
        {yatskar2016stating}
\bibfield{author}{\bibinfo{person}{Mark Yatskar}, \bibinfo{person}{Vicente Ordonez}, {and} \bibinfo{person}{Ali Farhadi}.} \bibinfo{year}{2016}\natexlab{}.
\newblock \showarticletitle{{Stating the obvious: Extracting visual common sense knowledge}}. In \bibinfo{booktitle}{\emph{Proceedings of the 2016 Conference of the North American Chapter of the Association for Computational Linguistics: Human Language Technologies}}. \bibinfo{pages}{193--198}.
\newblock


\bibitem[Yu and Riedl(2012)]%
        {yu2012sequential}
\bibfield{author}{\bibinfo{person}{Hong Yu} {and} \bibinfo{person}{Mark~O Riedl}.} \bibinfo{year}{2012}\natexlab{}.
\newblock \showarticletitle{A sequential recommendation approach for interactive personalized story generation.}. In \bibinfo{booktitle}{\emph{AAMAS}}, Vol.~\bibinfo{volume}{12}. \bibinfo{pages}{71--78}.
\newblock


\bibitem[Yu et~al\mbox{.}(2017)]%
        {yu2017hierarchically}
\bibfield{author}{\bibinfo{person}{Licheng Yu}, \bibinfo{person}{Mohit Bansal}, {and} \bibinfo{person}{Tamara~L Berg}.} \bibinfo{year}{2017}\natexlab{}.
\newblock \showarticletitle{Hierarchically-Attentive RNN for Album Summarization and Storytelling}. In \bibinfo{booktitle}{\emph{Proc. of the Conference on Empirical Methods for Natural Language Processing}}.
\newblock


\bibitem[Zarei et~al\mbox{.}(2020)]%
        {10.1145/3313831.3376331}
\bibfield{author}{\bibinfo{person}{Niloofar Zarei}, \bibinfo{person}{Sharon~Lynn Chu}, \bibinfo{person}{Francis Quek}, \bibinfo{person}{Nanjie~'Jimmy' Rao}, {and} \bibinfo{person}{Sarah~Anne Brown}.} \bibinfo{year}{2020}\natexlab{}.
\newblock \showarticletitle{Investigating the Effects of Self-Avatars and Story-Relevant Avatars on Children's Creative Storytelling}. In \bibinfo{booktitle}{\emph{Proceedings of the 2020 CHI Conference on Human Factors in Computing Systems}} (Honolulu, HI, USA) \emph{(\bibinfo{series}{CHI '20})}. \bibinfo{publisher}{Association for Computing Machinery}, \bibinfo{address}{New York, NY, USA}, \bibinfo{pages}{1–11}.
\newblock
\showISBNx{9781450367080}
\urldef\tempurl%
\url{https://doi.org/10.1145/3313831.3376331}
\showDOI{\tempurl}


\bibitem[Zhang et~al\mbox{.}(2021)]%
        {10.1145/3411763.3451785}
\bibfield{author}{\bibinfo{person}{Chao Zhang}, \bibinfo{person}{Cheng Yao}, \bibinfo{person}{Jianhui Liu}, \bibinfo{person}{Zili Zhou}, \bibinfo{person}{Weilin Zhang}, \bibinfo{person}{Lijuan Liu}, \bibinfo{person}{Fangtian Ying}, \bibinfo{person}{Yijun Zhao}, {and} \bibinfo{person}{Guanyun Wang}.} \bibinfo{year}{2021}\natexlab{}.
\newblock \showarticletitle{StoryDrawer: A Co-Creative Agent Supporting Children's Storytelling through Collaborative Drawing}. In \bibinfo{booktitle}{\emph{Extended Abstracts of the 2021 CHI Conference on Human Factors in Computing Systems}} (Yokohama, Japan) \emph{(\bibinfo{series}{CHI EA '21})}. \bibinfo{publisher}{Association for Computing Machinery}, \bibinfo{address}{New York, NY, USA}, Article \bibinfo{articleno}{354}, \bibinfo{numpages}{6}~pages.
\newblock
\showISBNx{9781450380959}
\urldef\tempurl%
\url{https://doi.org/10.1145/3411763.3451785}
\showDOI{\tempurl}


\bibitem[Zhang et~al\mbox{.}(2022)]%
        {10.1145/3491102.3501914}
\bibfield{author}{\bibinfo{person}{Chao Zhang}, \bibinfo{person}{Cheng Yao}, \bibinfo{person}{Jiayi Wu}, \bibinfo{person}{Weijia Lin}, \bibinfo{person}{Lijuan Liu}, \bibinfo{person}{Ge Yan}, {and} \bibinfo{person}{Fangtian Ying}.} \bibinfo{year}{2022}\natexlab{}.
\newblock \showarticletitle{StoryDrawer: A Child–AI Collaborative Drawing System to Support Children's Creative Visual Storytelling}. In \bibinfo{booktitle}{\emph{Proceedings of the 2022 CHI Conference on Human Factors in Computing Systems}} (New Orleans, LA, USA) \emph{(\bibinfo{series}{CHI '22})}. \bibinfo{publisher}{Association for Computing Machinery}, \bibinfo{address}{New York, NY, USA}, Article \bibinfo{articleno}{311}, \bibinfo{numpages}{15}~pages.
\newblock
\showISBNx{9781450391573}
\urldef\tempurl%
\url{https://doi.org/10.1145/3491102.3501914}
\showDOI{\tempurl}


\bibitem[Zook et~al\mbox{.}(2012)]%
        {zook2012automated}
\bibfield{author}{\bibinfo{person}{Alexander Zook}, \bibinfo{person}{Stephen Lee-Urban}, \bibinfo{person}{Mark~O Riedl}, \bibinfo{person}{Heather~K Holden}, \bibinfo{person}{Robert~A Sottilare}, {and} \bibinfo{person}{Keith~W Brawner}.} \bibinfo{year}{2012}\natexlab{}.
\newblock \showarticletitle{Automated scenario generation: toward tailored and optimized military training in virtual environments}. In \bibinfo{booktitle}{\emph{Proceedings of the international conference on the foundations of digital games}}. \bibinfo{pages}{164--171}.
\newblock


\end{thebibliography}

\pagebreak

\appendix
\section{Writing Interface}
\label{appendix}

\begin{figure}[ht!]
    \centering
    %\vspace{-1in}
    \includegraphics[width=0.45\textwidth]{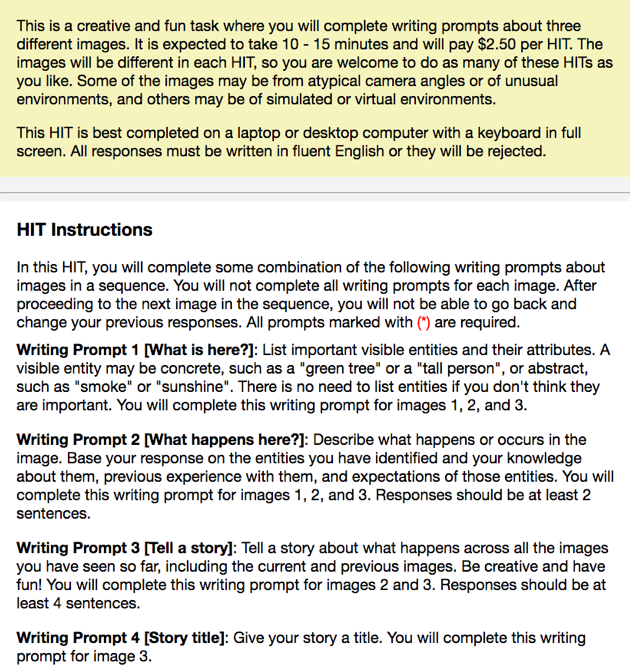}
    \caption{Instructions given prior to accepting the AMT HIT}
    \label{fig:hit-instr}
\end{figure}

\begin{figure*}[ht]
    \centering
    \includegraphics[width=0.95\textwidth]{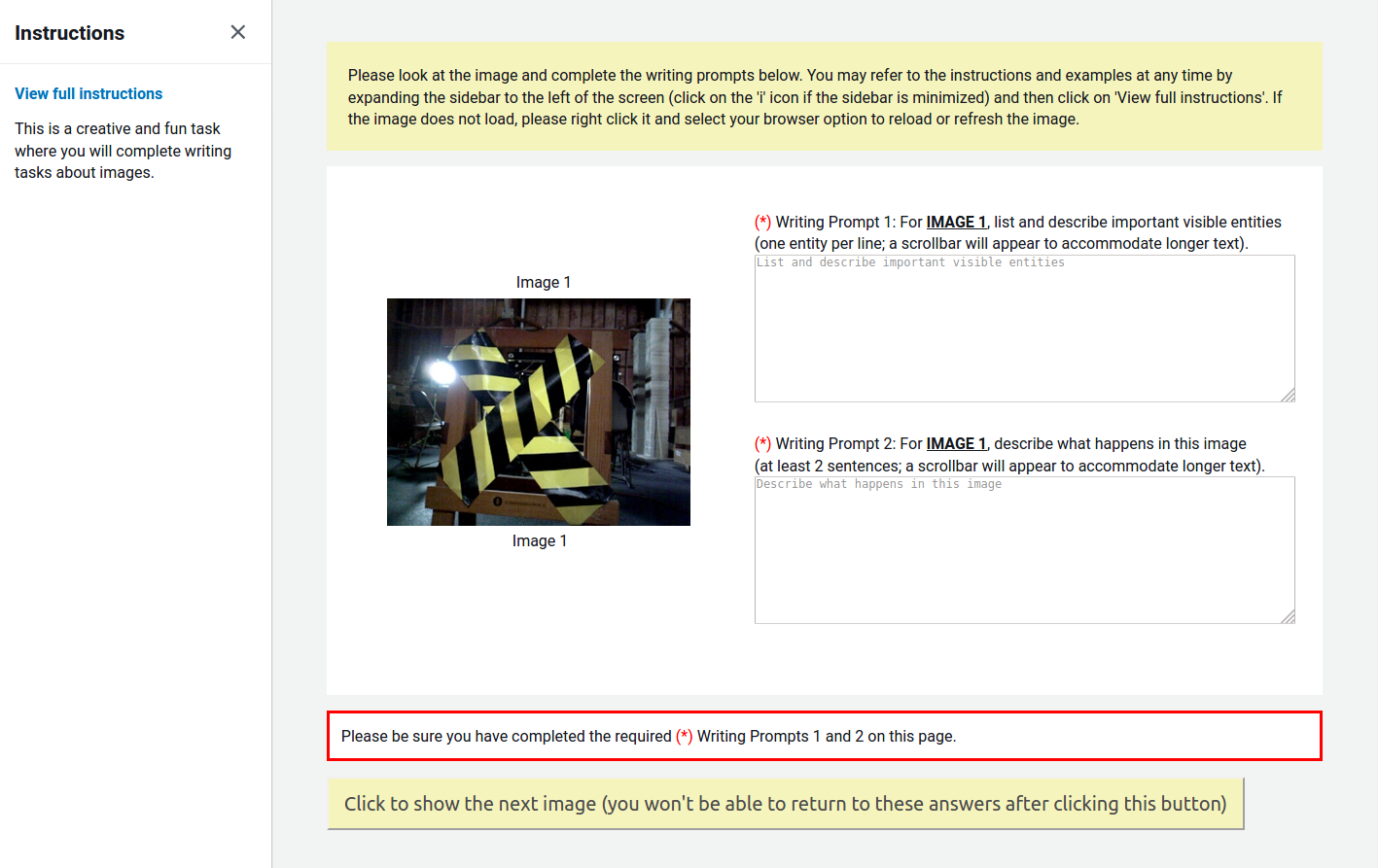}
    \caption{First page of the AMT data collection}
    \label{fig:hit-page1}
\end{figure*}

\begin{figure*}[ht]
    \centering
    \includegraphics[width=0.95\textwidth]{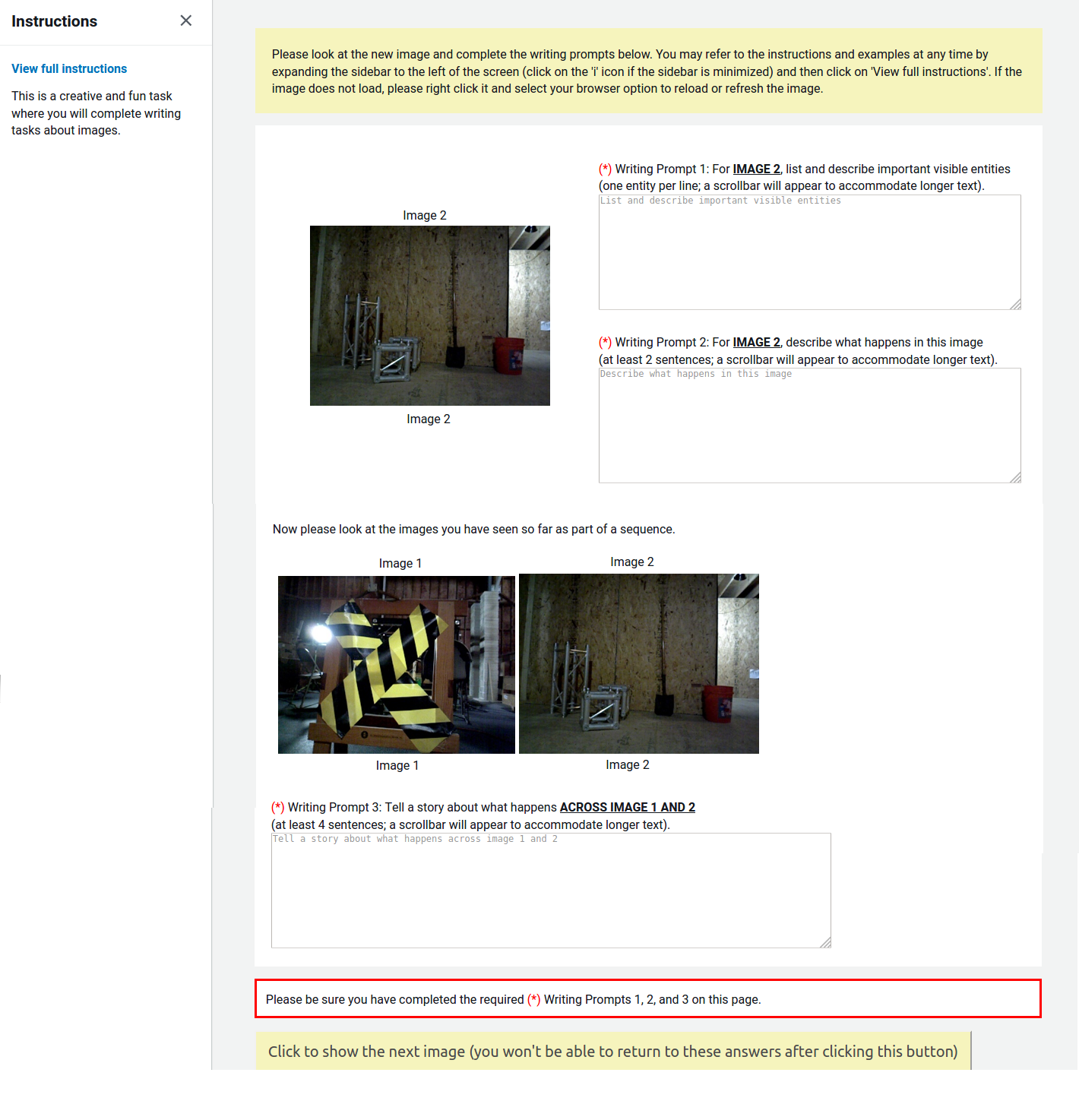}
    \caption{Second page of the AMT data collection}
    \label{fig:hit-page2}
\end{figure*}

\begin{figure*}[ht]
    \centering
    \includegraphics[width=0.95\textwidth]{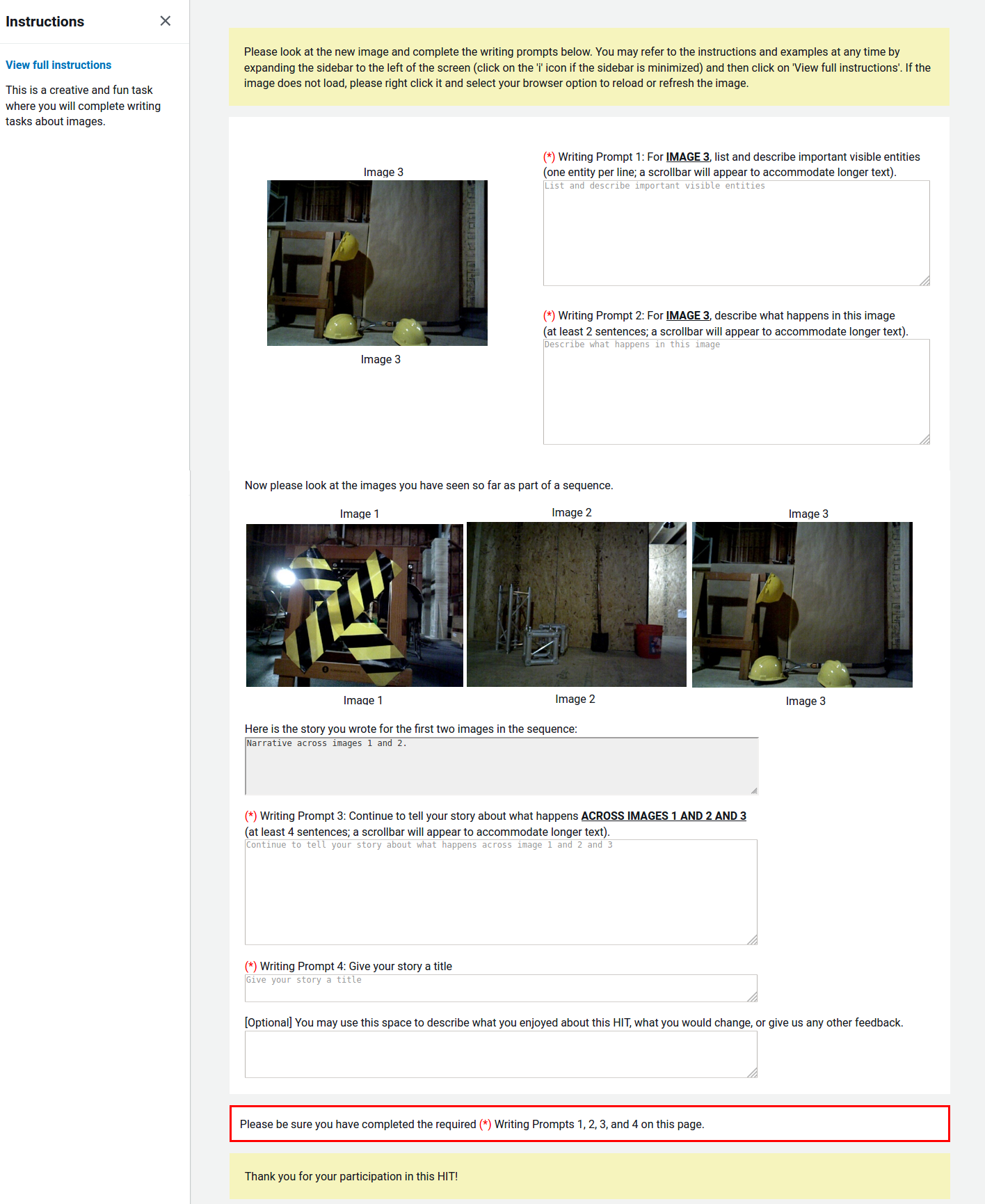}
    \caption{Third page of the AMT data collection}
    \label{fig:hit-page3}
\end{figure*}

Figure~\ref{fig:hit-instr} is a screenshot of the instructions shown to authors prior to them accepting the HIT. Figures~\ref{fig:hit-page1}--\ref{fig:hit-page3} are screenshots of the AMT webpage for the data collection of a SAR Image-set. New images and facets are revealed on each webpage. Here, the phrase ``writing prompt'' is synonymous with ``facet,'' where writing prompts 1, 2, 3 and 4 correspond with the Entity, Scene, Narrative, and Title Facet respectively.

\pagebreak

\end{document}